\documentclass{article}

    \PassOptionsToPackage{numbers, compress}{natbib}

\usepackage[preprint]{neurips_2024}

\usepackage[utf8]{inputenc} 
\usepackage[T1]{fontenc}    
\usepackage{xcolor} 
\definecolor{darkblue}{rgb}{0, 0.17, 0.55}
\definecolor{darkgreen}{rgb}{0, 0.55, 0.12}
\definecolor{darkred}{rgb}{0.6,0,0}
\definecolor{darkgreen}{rgb}{0,0.6,0}
\usepackage[hidelinks,
              colorlinks,
              urlcolor  = darkblue,
              citecolor = darkblue,]{hyperref}    
\usepackage{url}            
\usepackage{booktabs}       
\usepackage{amsfonts}       
\usepackage{nicefrac}       
\usepackage{microtype}      
\usepackage{xcolor}         
\usepackage{graphicx}
\usepackage{subfigure}
\usepackage{booktabs}
\newcommand{\red}[1]{\textcolor{red!80!black}{$(+#1)$}}
\newcommand{\green}[1]{\textcolor{green!50!black}{$(-#1)$}}

\usepackage{subcaption}

\usepackage{algorithm}
\usepackage{algorithmic}
\usepackage{colortbl}
\usepackage{wrapfig}

\newcommand{\distdiff}{\text{DistDiff}}
\newcommand{\caltech}{\text{Caltech-101}}
\newcommand{\cars}{\text{StanfordCars}}

\newcommand{\pathmnist}{\text{PathMNIST}}

\usepackage{amsmath,amsfonts,bm}

\def\eqref#1{equation~\ref{#1}}

\def\1{\bm{1}}

\def\vtheta{{\bm{\theta}}}

\def\vb{{\bm{b}}}
\def\vc{{\bm{c}}}

\def\ve{{\bm{e}}}

\def\vg{{\bm{g}}}

\def\vp{{\bm{p}}}

\def\vx{{\bm{x}}}
\def\vy{{\bm{y}}}
\def\vz{{\bm{z}}}

\DeclareMathAlphabet{\mathsfit}{\encodingdefault}{\sfdefault}{m}{sl}
\SetMathAlphabet{\mathsfit}{bold}{\encodingdefault}{\sfdefault}{bx}{n}

\def\gD{{\mathcal{D}}}

\def\gG{{\mathcal{G}}}

\def\gN{{\mathcal{N}}}

\def\gP{{\mathcal{P}}}

\def\gU{{\mathcal{U}}}

\title{Distribution-Aware Data Expansion with \\Diffusion Models}

\author{%
  Haowei Zhu$^*$ \\
  Tsinghua University \\
  \And
  Ling Yang \thanks{Equal Contribution.}\\
  Peking University \\
  \And
  Jun-Hai Yong \\
  Tsinghua University \\
  \And
  Hongzhi Yin\\
  University of Queensland
  \And
  Jiawei Jiang\\
  Wuhan University
  \And
  Meng Xiao \\
  CNIC, CAS \\
  \And
  Wentao Zhang\\
  Peking University\\
  \texttt{wentao.zhang@pku.edu.cn}
  \And
  Bin Wang \\
  Tsinghua University \\
  \texttt{wangbins@tsinghua.edu.cn} \\
}

\begin{document}

\maketitle

\begin{abstract}
The scale and quality of a dataset significantly impact the performance of deep models. However, acquiring large-scale annotated datasets is both a costly and time-consuming endeavor. To address this challenge, dataset expansion technologies aim to automatically augment datasets, unlocking the full potential of deep models. Current data expansion techniques include image transformation and image synthesis methods. Transformation-based methods introduce only local variations, leading to limited diversity. In contrast, synthesis-based methods generate entirely new content, greatly enhancing informativeness. However, existing synthesis methods carry the risk of distribution deviations, potentially degrading model performance with out-of-distribution samples. In this paper, we propose \textbf{DistDiff}, a training-free data expansion framework based on the \textbf{dist}ribution-aware \textbf{diff}usion model. DistDiff constructs hierarchical prototypes to approximate the real data distribution, optimizing latent data points within diffusion models with hierarchical energy guidance. We demonstrate its capability to generate distribution-consistent samples, significantly improving data expansion tasks. DistDiff consistently enhances accuracy across a diverse range of datasets compared to models trained solely on original data. Furthermore, our approach consistently outperforms existing synthesis-based techniques and demonstrates compatibility with widely adopted transformation-based augmentation methods. Additionally, the expanded dataset exhibits robustness across various architectural frameworks. Our code is available at \href{https://github.com/haoweiz23/DistDiff}{https://github.com/haoweiz23/DistDiff}.
\vspace{-10pt}
\end{abstract}

\section{Introduction}
\label{Intro}
A substantial number of training samples are essential for unlocking the full potential of deep networks. However, the manual collection and labeling of large-scale datasets are both costly and time-intensive. This makes it difficult to expand data-scarce datasets. Therefore, it is of great value to study how to expand high-quality training data in an efficient and scalable way \cite{Cherti_2023_CVPR}.

Automatic data expansion technology can alleviate the data scarcity problem by augmenting or creating diverse samples, it mitigates the bottleneck associated with limited data, thereby improving model's downstream performance and fostering greater generalization \cite{dunlap2023diversify, gifsd}. One simple yet effective strategy is employing image transformation techniques such as cropping, rotation, and erasing to augment samples \cite{shorten2019survey}. Although these methods prove effective and have been widely applied in various fields, their pre-defined perturbations only introduce local variations to the images, thereby falling short in providing a diverse range of content change. 
In recent times, generative models have gained considerable attention \cite{gan, glide, DALLE2, stablediffusion, imagen, parmar2023zero}, exhibit impressive performance in various areas like image inpainting \cite{lugmayr2022repaint, Palette}, super-resolution \cite{sr, sr2}, and video generation \cite{ImagenVideo, Mei_Patel_2023}. Generative models leverage text and image conditions to create images with entirely novel content, harnessing the expansive potential of data expansion \cite{dhariwal2021diffusion}. 
Nevertheless, there is a risk of generating images that deviate from the real data distribution. 
\begin{wrapfigure}{r}{0.5\textwidth}
\vspace{-10px}
\centering
\centerline{\includegraphics[width=0.9\linewidth]{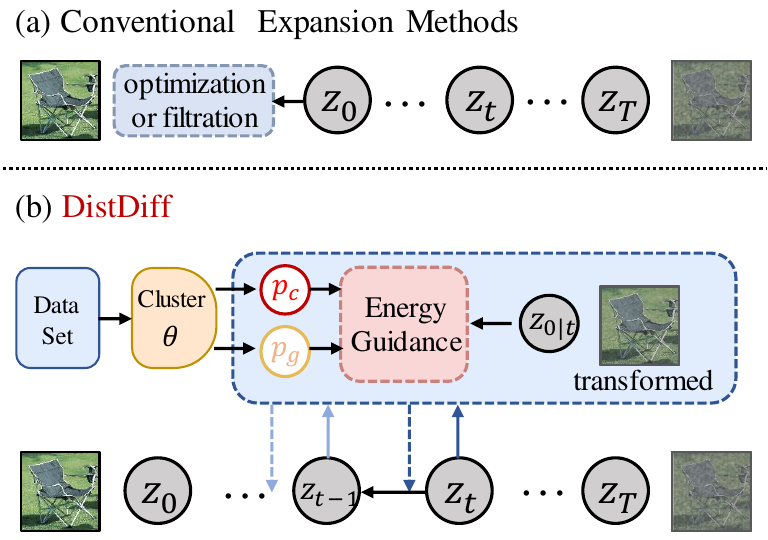}}
\caption{A comparison unveils distinctions between conventional data expansion methods and our innovative distribution-aware diffusion framework, benefiting from hierarchical clustering and multi-step energy guidance.}
\label{fig:concept}
\vspace{-10px}
\end{wrapfigure}
Therefore, when employing diffusion models for dataset expansion tasks, further research is necessary to ensure a match between synthetic data distributions and real data distributions.

There are several strategies aimed at mitigating the risk of distribution shift, which can be broadly categorized into two groups: training-based and training-free methods. The training-based methods \cite{moon2022fine, ruiz2023dreambooth, xie2023difffit} fine-tune pre-trained diffusion models to adapt target dataset, necessitating additional training costs and increasing the likelihood of overfitting on small-scale datasets. Other training-free methods \cite{feng2023diverse, he2022synthetic} eliminate potentially noisy samples by designing optimizing and filtering strategies, but they still struggle to generate data that conforms to the real data distribution.

In this work, we propose a training-free data expansion framework, dubbed \textbf{Dist}ribution-Aware \textbf{Diff}usion (DistDiff) to optimize generation results. As shown in Figure \ref{fig:concept}, DistDiff initially approximates the true data distribution using class-level and group-level prototypes obtained through hierarchical clustering. Subsequently, DistDiff utilizes these prototypes to formulate two synergistic energy functions. A residual multiplicative transformation is then applied to the latent data points, enabling the generation of data distinct from the original. Following this, the hierarchical energy guidance process refines intermediate predicted data points, optimizing the diffusion model to generate data samples that are consistent with the underlying distribution. DistDiff ensures fidelity and diversity in the generated samples through distribution-aware energy guidance.
Experimental results demonstrate that DistDiff outperforms advanced data expansion techniques, producing better expansion effects and significantly improving downstream model performance. 
Our contributions can be summarized as follows:
\begin{itemize}
\item We introduce a novel diffusion-based data expansion algorithm, named DistDiff, which facilitates distribution-consistent data augmentation without requiring re-training. 
\item By leveraging hierarchical prototypes to approximate data distribution, we propose an effective distribution-aware energy guidance at both class and group levels in the diffusion sampling process. 
\item The experimental results illustrate that our DistDiff is capable of generating high-quality samples, surpassing existing image transformation and synthesis methods significantly.
\vspace{-5pt}
\end{itemize}

\section{Related Work}
\vspace{-5pt}
\subsection{Transformation-Based Data Augmentation}
Traditional data augmentation techniques \cite{chen2020gridmask, cubuk2020randaugment, hendrycks2019augmix, yun2019cutmix, zhang2017mixup, zhao2020maximum, cutout} typically involve expanding the dataset through distortive transformations, aiming to enhance the model's ability to capture data invariance and mitigate overfitting \cite{shorten2019survey}. For instance, scale invariance is cultivated through random cropping and scaling, while rotation invariance is developed through random rotation and flipping. 
Region mask-based methods \cite{chen2020gridmask, cutout_2017, cutout} enhance model robustness against target occlusion by strategically obscuring portions of the target area. Interpolation-based methods \cite{hendrycks2019augmix, yun2019cutmix, zhang2017mixup} generate virtual samples by randomly blending content from two images. RandAugment \cite{cubuk2020randaugment} further boosts augmentation effectiveness by sampling from a diverse range of augmentation strategies. However, these methods induce only subtle changes on the original data through transformation, deletion, and blending, leading to a lack of diversity. Moreover, they are predefined and uniformly applied across the entire dataset, which may not be optimal for varying data types or scenarios.

\subsection{Synthesis-Based Data Augmentation}
Generative data augmentation aims to leverage generative models to approximate the real data distribution, generating samples with novel content to enhance data diversity. GAN \cite{gan} excels at learning data distributions and producing unseen samples in an unsupervised manner \cite{antoniou2017data, gurumurthy2017deligan, li2022bigdatasetgan, mariani2018bagan, zhang2021datasetgan, zhao2022synthesizing, xu2023handsoff}. 
While their efficacy has been demonstrated across diverse downstream tasks, studies indicate that training existing models like ResNet50 \cite{resnet} on images synthesized by BigGAN \cite{brock2018large} yields subpar results compared to training on real images. This disparity in performance can be attributed to the limited diversity and potential domain gap between synthesized samples and real images. Additionally, the training processes of GAN are notoriously unstable, particularly with a low-data regime, and suffer from mode collapse, resulting in a lack of diversity \cite{bansal2023leaving, gowal2021improving, ravuri2019classification}. 
In contrast, diffusion model-based methods \cite{yang2023diffusion} can offer better controllability and superior customization capabilities. Text-to-image models such as Stable Diffusion \cite{stablediffusion}, DALL-E 2 \cite{DALLE2} and RPG \cite{yang2024mastering} have demonstrated the creation of compelling high-resolution images \cite{yang2024improving, yang2024crossmodal,zhang2024realcompo}. Recently, large-scale text-to-image models have been used for data generation \cite{azizi2023synthetic, dunlap2024diversify, li2023guiding, shipard2023diversity, wu2024datasetdm, wu2023diffumask}. For example, LECF~\cite{he2022synthetic} utilized GLIDE \cite{glide} to generate images, filtering low-confidence samples to enhance zero-shot and few-shot image classification performance. SGID \cite{li2023semantic} leverages image descriptions generated by BLIP~\citep{li2022blip} to enhance the semantic consistency of generated samples. Feng et al.~\cite{feng2023diverse} filters out low-quality samples based on feature similarity between generated and reference images. GIF \cite{gifsd} creates new informative samples through prediction entropy and feature divergence optimization. However, it's crucial to note that datasets generated by existing methods may exhibit distribution shifts, impacting image classification performance significantly. Zhou et al. \cite{zhou2023training} address this issue by employing diffusion inversion to mitigate distributional shifts. In contrast, we propose a training-free approach, leveraging hierarchical prototypes as optimization targets to guide the generation process, thereby addressing distributional shifts. This approach offers the advantage of avoiding additional computational costs and overfitting issues associated with fine-tuning diffusion models.

\begin{figure*}[t]
\begin{center}
\centerline{\includegraphics[width=0.9\linewidth]{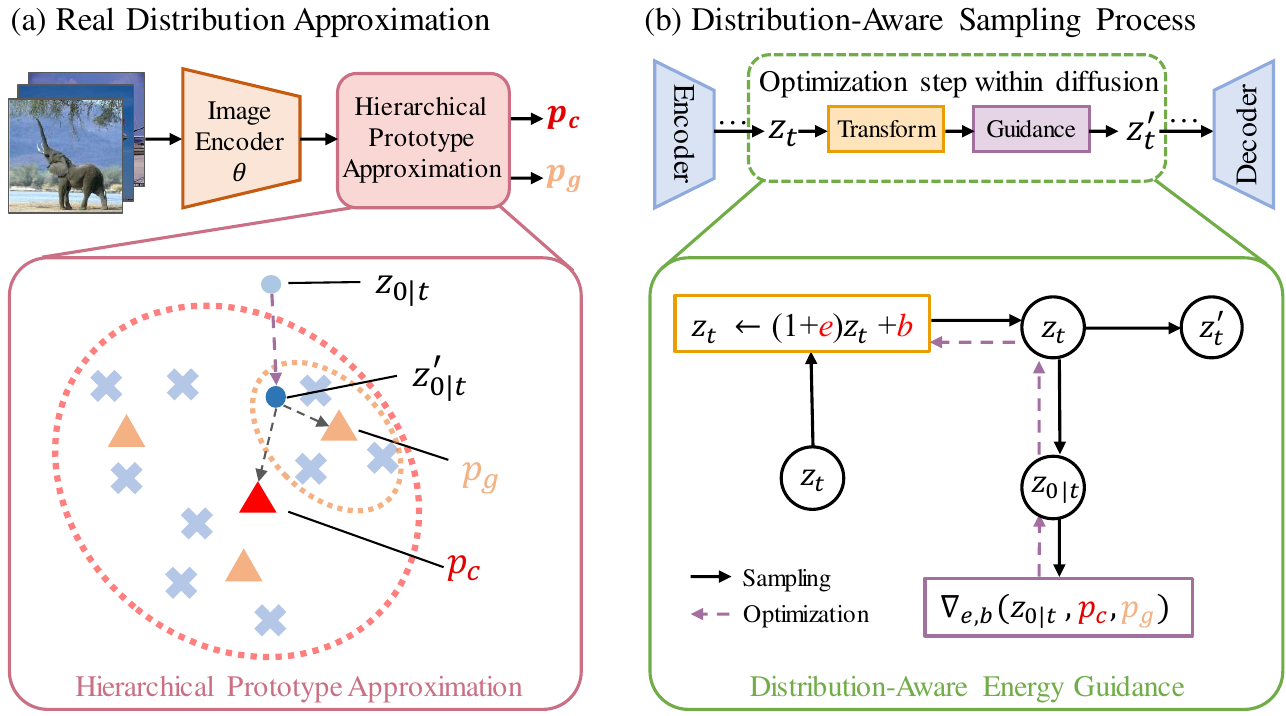}}
\caption{Overview of the DistDiff pipeline. DistDiff enhances the generation process in diffusion models with distribution-aware optimization. It approximates the real data distribution using hierarchical prototypes  $\vp_c$ and $\vp_g$, optimizing the sampling process through distribution-aware energy guidance. Subsequently, original generated data point $\vz_t$ is refined for improved alignment with the real distribution.}
\label{fig:overview}
\end{center}
\vspace{-20pt}
\end{figure*}

\vspace{-5pt}
\section{Method}
\label{sec:method}
In this study, we introduce a distribution-aware data expansion framework utilizing Stable Diffusion as a prior generation model. This framework guides the diffusion model based on hierarchical prototype guidance criteria. As illustrated in Figure \ref{fig:overview}, our DistDiff initially employs an image encoder $\theta$ to extract instance features and subsequently derives hierarchical prototypes to approximate the real data distribution. Next, for a given seed image and corresponding text prompts, we extract image's latent feature and apply stochastic noise to it. Subsequently, in the denoising process, we optimize the latent features using a training-free hierarchical energy guidance process. Our optimization strategy ensures that the generated samples not only match the distribution but also carry new information to enhance model training.

\subsection{Task Definition} 
In the context of a small-scale training dataset, the data expansion task is designed to augment the original dataset $\gD_o=\{\vx_i, \vy_i\}_{i=1}^{n_o}$ with a new set of synthetic samples, referred to as $\gD_s=\{\vx'_i, \vy'_i\}_{i=1}^{n_s}$. Here $\vx_i$ and $\vy_i$ represent the sample and its corresponding label, where $n_o$ and $n_s$ respectively denote the original sample quantity and the synthetic sample quantity. The objective is to enhance the performance of a deep learning model trained on both the original dataset $\gD_o$ and the expanded dataset $\gD_s$ compared to a model trained solely on the original $\gD_o$. The crucial aspect lies in ensuring that the generated dataset is highly consistent with the distribution of the original dataset while being as informative as possible.

\subsection{Hierarchical Prototypes Approximate Data Distribution}
\label{sec:hierarchical_design}
Prototypes have been widely employed in class incremental learning methods to retain information about each class \cite{masana2022class, rebuffi2017icarl}. In this work, we propose two levels of prototypes to capture the original data distribution. Firstly, the class-level prototypes $\vp_\mathrm{c}$ are obtained by averaging feature vectors within the same class. The class vector aggregates high-level statistical information to characterize all samples from the same class as a collective entity. However, as class-level prototypes represent the class feature space as a single vector, potentially reducing informativeness, we further introduce group-level prototypes to capture the structure of the class feature space. Specifically, we divide all samples from the same class into $K$ groups using agglomerative hierarchical clustering, followed by averaging feature vectors within each group to obtain $K$ group prototypes $\vp_\mathrm{g}=\{\vp_\mathrm{g}^1, \vp_\mathrm{g}^2, ..., \vp_\mathrm{g}^K\}$. Instances with similar patterns are grouped together. Transitioning from class-level to group-level, the prototypes encapsulate abstract distribution information of the class at different scales.

Thanks to these hierarchical prototypes, we design two function $\gD_\vtheta^\mathrm{c}$ and $\gD_\vtheta^\mathrm{g}$ to evaluate the degree of distribution matching:
\begin{equation}\label{Eq:s_class}
    \gD_\vtheta^\mathrm{c}(\vx, \vp_\mathrm{c}) = \lVert \vtheta(\vx) - \vp_\mathrm{c} \rVert_2,
\end{equation}
\begin{equation}\label{Eq:herichical_p}
\begin{gathered}
    \gD_\vtheta^\mathrm{g}(\vx, \vp_\mathrm{g}) = \lVert \vtheta(\vx) - \vp_\mathrm{g}^j \rVert_2,\\
    \mathrm{ s.t. } j = \arg\max(\mathrm{cos}(\vtheta(\vx), \{\vp_\mathrm{g}^j\}_{j=1}^K)),
\end{gathered}
\end{equation}
where $\vtheta(\cdot)$ means feature extractor, which could be ResNet \cite{resnet}, CLIP \cite{CLIP} or other deep models.

Note that these two functions evaluate the score of distribution matching from two perspectives. The value will be lower when $\vx$ is more consistent with the real distribution. As shown in Figure \ref{fig:overview} (a), $\gD_\vtheta^\mathrm{c}$ gauges the distance of sample features to the class center, resulting in low scores for easy samples while high scores for hard samples that are situated at the boundaries of the distribution. On the other hand, $\gD_\vtheta^\mathrm{g}$ assesses distance from the group-level, offering lower scores for hard samples, while still maintaining relatively high scores for outlier samples. Consequently, these two scores mutually reinforce each other and are indispensable.

\subsection{Transform Data Points} 
Given a reference sample $(\vx, \vy)$, the pre-trained large-scale diffusion model $\gG$ can generate new samples $\vx'$ with novel content. We formalize this process as $\vx' = \gG(\Phi(\vx) + \delta)$, where $\Phi(\vx)$ represents the latent feature representation and $\delta$ is the perturbation applied to latent features. Drawing inspiration from GIF \cite{gifsd}, we introduce residual multiplicative transformation to the latent feature $\vz=\Phi(\vx)$ using randomly initialized channel-level noise $\ve\sim\gU(0,1)$ and $\vb\sim\gN(0,1)$. We impose an $\epsilon$-ball constraint on the transformed feature to control the degree of adjustment within a reasonable range, \textit{i.e.}, $\lVert \vz - \tilde{\vz} \rVert_\infty \leq \epsilon$, and derive $\tilde{\vz}$ as follows. 
\begin{equation}\label{Eq:channelnoise}
    \tilde{\vz}=\gP_{z,\epsilon}(\tau(\vz)) = \gP_{z,\epsilon}((1+\ve)\vz+\vb),
\end{equation}
where $\tau(\cdot)$ represents the transformation function and  $\gP_{z,\epsilon}(\cdot)$  denotes the projection of the transformed feature $\tilde{\vz}$ onto the $\epsilon$-ball of the original latent feature $\vz$.

Now, the key challenge lies in optimizing $\ve$ and $\vb$ to create new samples that align closely with the real data distribution.
Another intuitive approach is to directly optimize latent features instead of performing residual multiplication transformations. However, directly optimizing latent features leads to minimal perturbations, making it challenging to achieve performance gains. We discuss this alternative approach in Section~\ref{sec:ab}.

\subsection{Distribution-Aware Diffusion Generation}
In a typical diffusion sampling process, the model iteratively predicts noise to progressively map the noisy $\vz_T$ into clean $\vz_0$. While existing data expansion methods \cite{feng2023diverse, he2022synthetic, li2023semantic} treat the generative model as a black box, focusing on filtering or optimizing the final generated $\vz_0$. The importance of the intermediate sampling stage is ignored, which plays a crucial role in ensuring data quality, especially as the image begins to take on a stable shape appearance. Diverging from prior approaches, we advocate for intervention at the intermediate denoising step for optimization.

Specifically, we first introduce energy guidance into standard reverse sampling process to optimize the transformation using Equation \ref{Eq:guidance}. As our energy guidance step is applied to the transformed data point, the transformed data point $\tilde{\vz_t}$ is denoted as $\vz_t$ for simplicity.

\begin{equation}\label{Eq:guidance}
\begin{aligned}
    \ve' = \ve - \rho \nabla_{\ve} \varepsilon(\vz_t, c), \\
    \vb' = \vb - \rho \nabla_{\vb} \varepsilon(\vz_t, c),
\end{aligned}
\end{equation}
where $\rho$ is the learning rate and $\varepsilon(\vz_t, \vc)$ is the energy function measuring the compatibility between the transformed noisy data point $\vz_t$ and the given condition $\vc$, representing the real data distribution in this work. Equation \ref{Eq:guidance} guides the sampling process and generates distribution-consistent samples. After that, the optimized $\vz'_{t}$ is obtained via Equation \ref{Eq:channelnoise}. However, directly measuring the distance between intermediate results $\vz_t$ with condition $\vc$ is impractical due to the difficulty in finding a pre-trained network that provides meaningful guidance when the input is noisy.

To address this issue, we leverage the capability that the diffusion model can predict the noise added to $\vz_t$, and thus predict a clean data point $\vz_{0|t}$, as shown in Equation \ref{Eq:z0}. Then, the new energy function $\gD_\vtheta(\vz_{0|t}, \vc)$ based on the predicted clean data point is constructed to approximate $\varepsilon(\vz_t, \vc)$.
\begin{equation}\label{Eq:z0}
    \vz_{0|t} = \frac{\vz_t-\sqrt{1-\alpha_t}\psi(\vz_t,t)}{\sqrt{\alpha_t}},
\end{equation}
where $\alpha_t$ represents the noise scale and $\psi$ is the learned denoising network.
Finally, we employ hierarchical prototypes $\vp_\mathrm{c}$ and $\vp_\mathrm{g}$ as conditions to construct our energy guidance in the following manner:
\setlength{\abovedisplayskip}{10pt}
\setlength{\belowdisplayskip}{10pt}
\begin{equation}\label{Eq:final_guidance}
\begin{aligned}
      \ve' = \ve - \rho \nabla_{e} (\gD_\vtheta^\mathrm{c}(\vz_{0|t}, \vp_\mathrm{c}) + \mathcal{D}_\vtheta^\mathrm{g}(\vz_{0|t}, \vp_\mathrm{g})), \\
      \vb' = \vb - \rho \nabla_{b} (\gD_\vtheta^\mathrm{c}(\vz_{0|t}, \vp_\mathrm{c}) + \mathcal{D}_\vtheta^\mathrm{g}(\vz_{0|t}, \vp_\mathrm{g})).
\end{aligned}
\end{equation}
Unlike existing methods that exclusively optimize the final sampling result $\vz_0$, our approach focuses on optimizing intermediate denoising steps within the sampling process. The detailed algorithm is shown in Appendix \ref{append:pseudocode}. This novel strategy leads to substantial improvements in optimization results and will be further explored in Section \ref{sec:ab}.

\section{Experiments}
\subsection{Experimental Setups}
\label{sec: experimental setup}
\paragraph{Datasets} We assess the performance of DistDiff across six image classification datasets, encompassing diverse tasks such as general object classification (Caltech-101 \cite{caltech101}, CIFAR100-Subset \cite{cifar}, ImageNette \cite{howardsmaller}), fine-grained classification (Cars \cite{car}), textual classification (DTD \cite{dtd}) and medical imaging (PathMNIST \cite{yang2023medmnist}). More details are provided in Appendix \ref{append:dataset}.

\paragraph{Compared Methods} We conduct a comparative analysis between DistDiff and conventional image transformation methods, as well as diffusion-based expansion methods. Traditional image transformation techniques considered in the comparison comprise AutoAugment \cite{cubuk2018autoaugment}, RandAugment~\cite{cubuk2020randaugment}, Random Erasing \cite{cutout},  GridMask \cite{chen2020gridmask}, and interpolation-based techniques like MixUp \cite{zhang2017mixup} and CutMix \cite{yun2019cutmix}. 
For generative-based methods, we include the direct application of stable diffusion for data expansion, as well as the most recent state-of-the-art method, Stable Diffusion 1.4 \cite{stablediffusion}, LECF \cite{he2022synthetic}, GIF-SD \cite{gifsd}. The implementation details of these techniques are provided in Appendix \ref{append:Synthesis-details} and \ref{append:Transformation-details}.

\subsection{Implementation Details}
In our experimental setup, we implement DistDiff based on Stable Diffusion 1.4 \cite{stablediffusion}. The images created by Stable Diffusion have a resolution of $512 \times 512$ for all datasets. 
Throughout the diffusion process, we employ the DDIM \cite{ddim} sampler for a $50$-step latent diffusion, with hyper-parameters for noise strength set at $0.5$ and classifier free guidance scale at $7.5$. The $\epsilon$ in Equation \ref{Eq:channelnoise} is $0.2$ by default. We use a ResNet-50 \cite{resnet} model trained from scratch on the original datasets as our guidance model. We assign $K=3$ to each class when constructing group-level prototypes, the learning rate $\rho$ is $10.0$, and optimization step $M$ is set to $20$ unless specified otherwise. 
After expansion, we concatenate the original dataset with synthetic data to create expanded datasets. We then train the classification model from random initialization for 100 epochs using these expanded datasets.
During model training, we process images through random cropping to $224 \times 224$ using random rotation and random horizontal flips. Our optimization strategy involves using the SGD optimizer with a momentum of $0.9$, and cosine decay with an initial learning rate of $0.1$. All results are averaged over three runs with different random seeds. More implementation details can be found in the Appendix \ref{append:implement}.


\subsection{Main Results}
\label{sec:main_results}
\paragraph{Comparison with Synthesis-Based Methods} 

\begin{wrapfigure}{r}{0.5\textwidth}
  \centering
  \vspace{-0.2in}
  \includegraphics[width=1.0\linewidth]{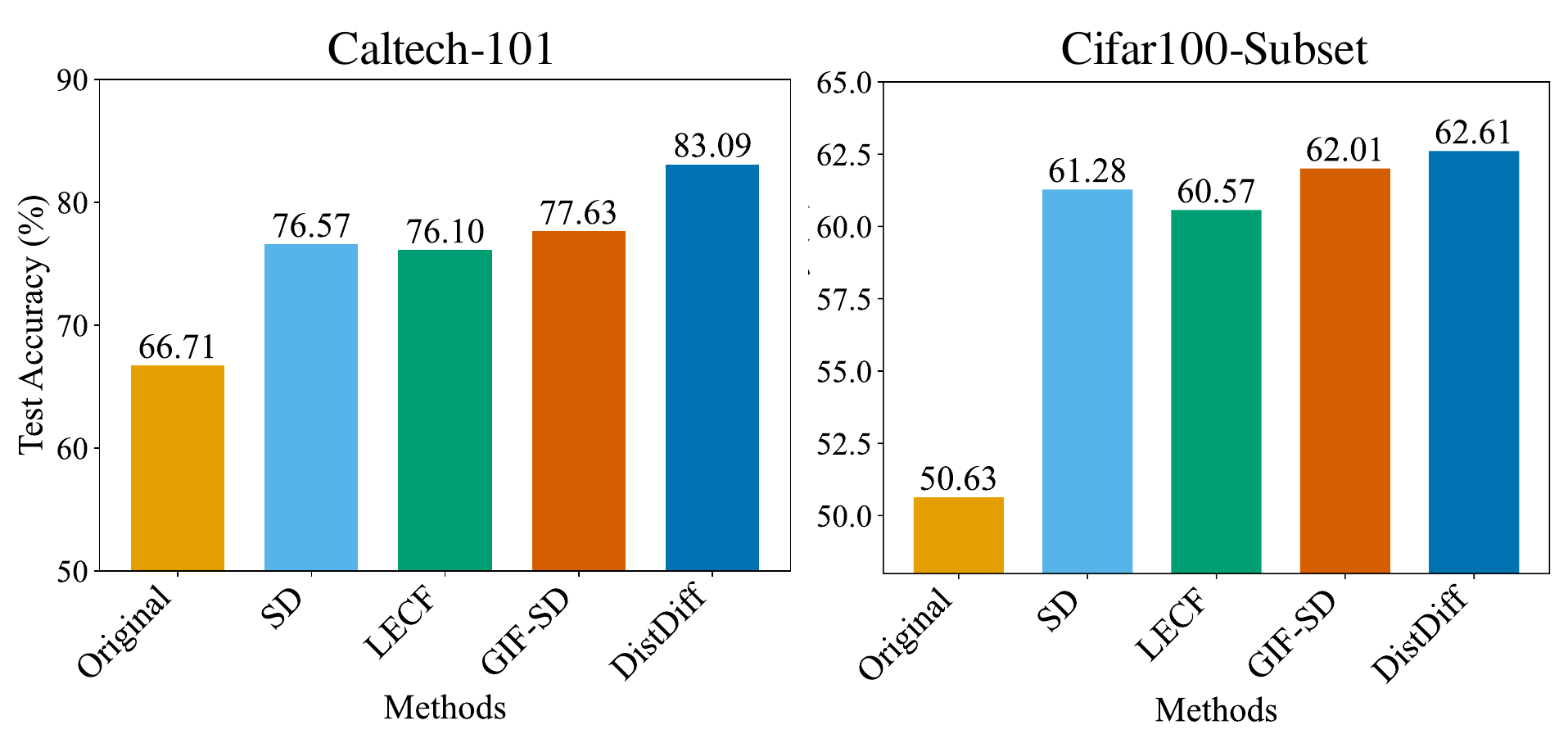}
  \vspace{-0.2in}
  \caption{\footnotesize{Our method outperforms state-of-the-art data expansion methods when trained on expanded datasets, underscoring the importance of a high-quality generator in training a classifier.}}
  \label{fig:sota_compare}
  \vspace{-0.2in}
\end{wrapfigure}

To investigate the effectiveness of methods for generating high-quality datasets for downstream classification model training, we initially compare our method, DistDiff, with existing synthesis-based methods on \caltech in terms of classification performance. Figure~\ref{fig:sota_compare} highlights the superiority of our method over state-of-the-art techniques. Compared to the original stable diffusion, DistDiff exhibits an average improvement of 6.25\%, illustrating that our method retains more distribution-aligned information from the original datasets. 
Additionally, GIF-SD \cite{gifsd}, which uses a pre-trained CLIP \cite{CLIP} model to enhance class-maintained information and employs KL-divergence to encourage batch-wise sample diversity, is also surpassed by our method by 5.46\% in accuracy. This can be attributed to DistDiff guiding the generation process from the distribution level with hierarchical prototypes, providing better optimization signals.

We also evaluated DistDiff against LECF \cite{he2022synthetic}, which enhances language prompts and filters samples with low confidence. 
We use LE enhanced to synthesize $5\times$ synthetic datasets and filter with different thresholds. We assessed multiple Clip Filtering strengths in LECF and found that LECF did not achieve better performance compared to the original Stable Diffusion. This is due to our SD baseline already generates high-quality samples, and the additional filtering post-process may lead to data loss.
Besides, both LECF and GIF-SD use auxiliary models to filter/guide the diffusion generation results. However, there are two main strengths of our DistDiff compared to these methods. First, DistDiff generates images end-to-end without requiring filter-based postprocessing. Second, our method is not sensitive to the classification performance of the auxiliary model, which is proved in the Appendix \ref{append:further_analysis}. This suggests that DistDiff is simpler and more generalizable for data expansion tasks. 
\vspace{-10pt}
\paragraph{Comparison with Transformation-Based Augmentation Methods}
In Table \ref{tab:augmentation}, we present a comparison of our methods with widely adopted data augmentation techniques for \caltech image classification. Our DistDiff method surpasses transformation-based augmentation methods by introducing a broader range of new content into images. Additionally, we demonstrate the compatibility of our approach with transformation-based data augmentation methods, leading to further improvements.

\begin{table}[htbp]
\centering
\caption{Comparison of transformation-based augmentation methods on \caltech. Our approach, combined with default augmentation (crop, flip, and rotate), consistently outperforms existing advanced transform-based methods and can be further improved by combining these techniques.}
\begin{tabular}{@{}lccccccc@{}}
\toprule
& Default & AutoAug & RandAug & Random Erasing & GridMask & MixUp & CutMix \\
\midrule
Original  & 66.71 & 74.34 & 74.07 & 74.22 & 73.88 & 78.64 & 70.13 \\
\distdiff & 83.38 & 82.93 & 83.21 & 83.05 & 83.48 & 81.06 & 85.27 \\
\bottomrule
\end{tabular}
\label{tab:augmentation}
\end{table}

\begin{figure*}[t]
\centering
\begin{minipage}[b]{0.24\textwidth}
    \centering
    \includegraphics[width=\textwidth]{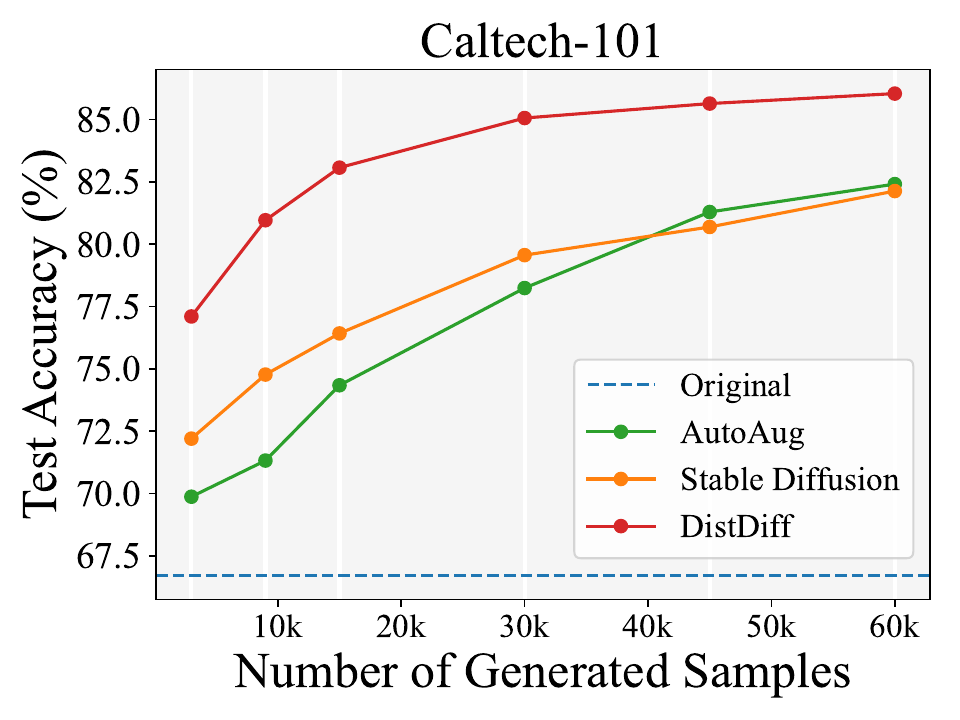}
\end{minipage}
\begin{minipage}[b]{0.24\textwidth}
    \centering
    \includegraphics[width=\textwidth]{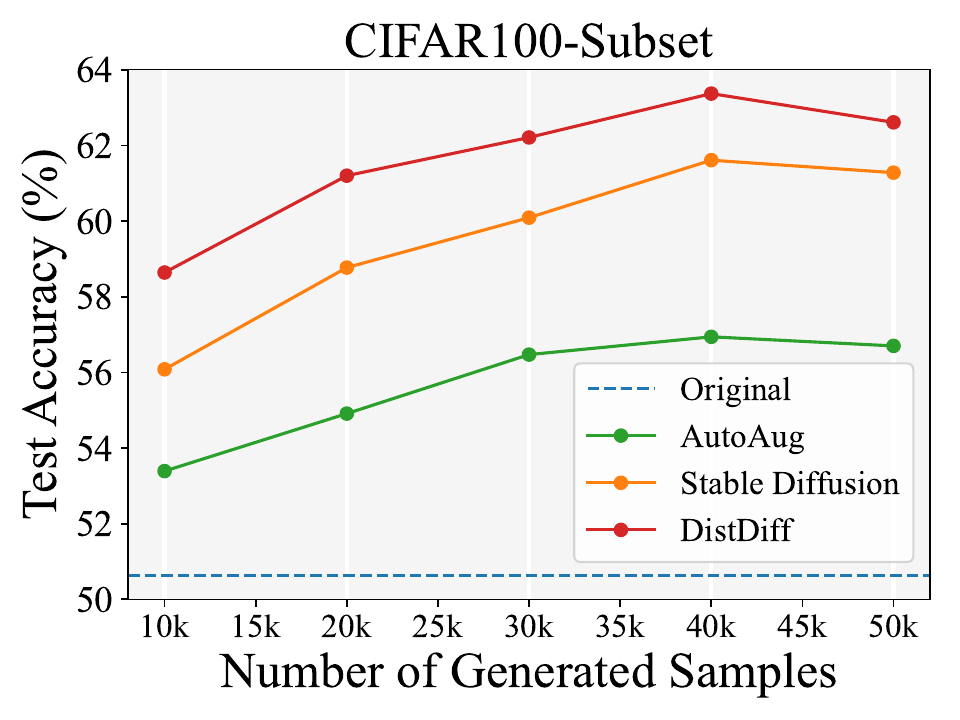}
\end{minipage}
\begin{minipage}[b]{0.24\textwidth}
    \centering
    \includegraphics[width=\textwidth]{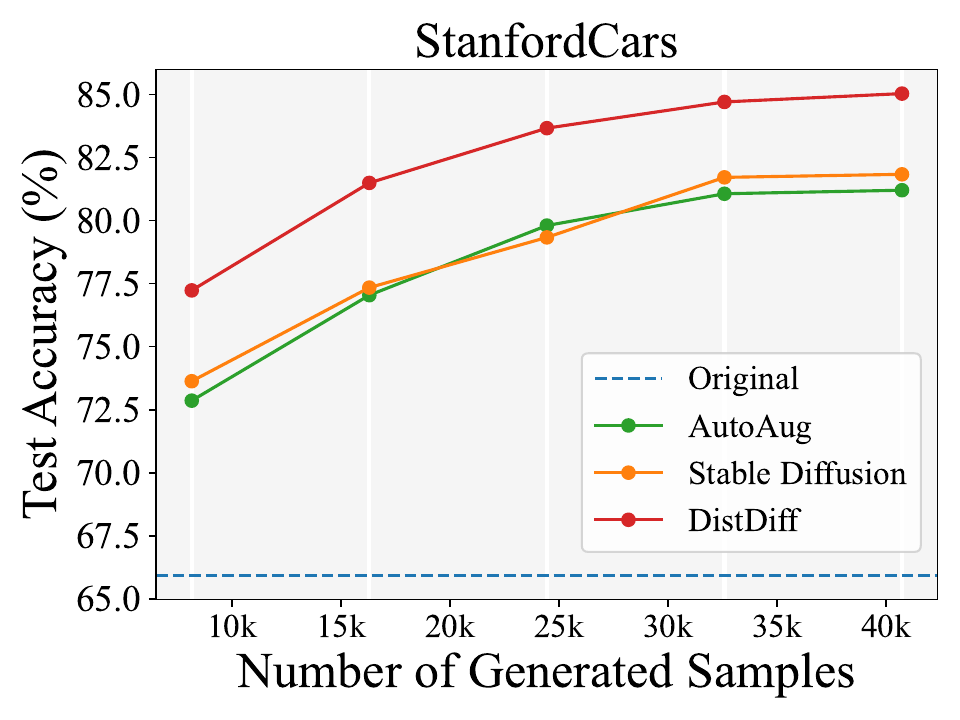}
\end{minipage}
\begin{minipage}[b]{0.24\textwidth}
    \centering
    \includegraphics[width=\textwidth]{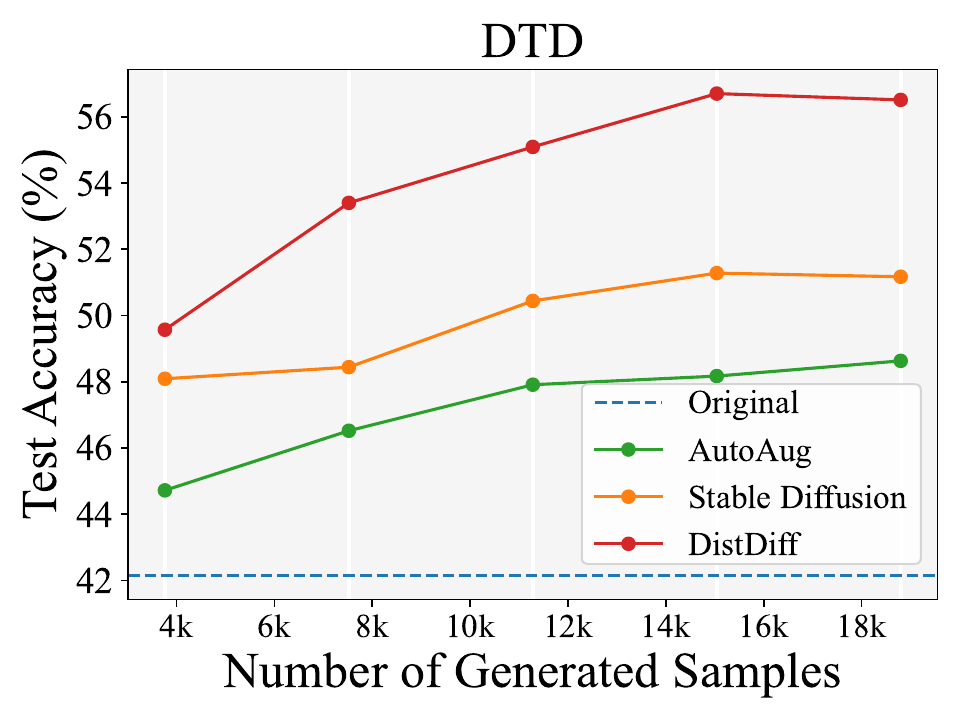}
\end{minipage}
\caption{\footnotesize{Performance comparison across different scale data sizes. Our method demonstrates significant improvements in classification model performance in both low-data and large-scale data scenarios, outperforming the transformation method AutoAug and the synthesized method Stable Diffusion 1.4.}}
\label{fig:scaling}
\vspace{-20pt}
\end{figure*}

\paragraph{Scaling in Number of Data} We evaluate the scalability of our approach by assessing its advantages in classification model training across four datasets.  We compare the performance of DistDiff with the original real dataset and strong augmentation method AutoAug \cite{cubuk2018autoaugment} with varying numbers of generated examples, as depicted in Figure \ref{fig:scaling}. As the data expansion scale increases, the corresponding improvement in accuracy also enlarges. The accuracy on \caltech achieved with our $5\times$ expansion surpasses even the $20\times$ expanded dataset obtained by AutoAug and diffusion baseline. This indicates that DistDiff exhibits superior efficiency in data expansion compared to existing methods.

\paragraph{Versatility to Various Architectures}  We conduct an in-depth assessment of the expanded datasets generated by \distdiff across four distinct backbones: ResNet-50 \cite{resnet}, ResNeXt-50 \cite{resnext}, WideResNet-50 \cite{wideresnet}, and MobileNetv2 \cite{mobilenetv2}. These backbones are trained from scratch on $5\times$ expanded \caltech dataset by \distdiff. The results presented in Table \ref{tab:model_comparison} affirm that our innovative methodology is effective and versatile across a spectrum of architectures.

\paragraph{Comparison with Stronger Classification Models} 
As we know, data expansion is typically applied in scenarios with data scarcity. However, if we use models pre-trained on large-scale datasets, the performance on the original training set can be significantly enhanced. In such cases, does our expanded dataset still offer improvements?
To validate our method, we fine-tuned a ResNet-50 model pre-trained on ImageNet-1k \cite{imagenet} for ImageNette, Caltech-101, and StanfordCars, and a LAION \cite{schuhmann2022laion} pre-trained CLIP-ViT-B32 \cite{Cherti_2023_CVPR} model for PathMNIST. As shown in Table \ref{table:pretrained}, the model achieved a high accuracy of 99.4\% on ImageNette, which is a subset of its pretrained datasets. Further data expansion resulted in a slight decrease in performance. Similarly, on the general image dataset \caltech, which shares significant overlap with ImageNet data, our method demonstrated only slight improvement.
However, on the more challenging fine-grained dataset \cars, our method demonstrated obvious 3.56\% accuracy improvement. For the medical image dataset \pathmnist, which exhibits a significantly different distribution, using DistDiff for data expansion effectively boosted classification performance by 5.18\%. This highlights the importance of scaling up data when transferring pre-trained models to downstream tasks that exhibit significant distribution shifts.

\begin{table}[h]
\centering
\caption{Comparison of using stronger pre-trained baseline models. On ImageNette \cite{imagenette} , Caltech-101 \cite{caltech101}, and StanfordCars \cite{car} datasets, we employ an ImageNet-1k \cite{imagenet} pre-trained ResNet-50 \cite{resnet} model. For the PathMNIST \cite{yang2023medmnist} dataset, we fine-tune using the stronger CLIP-ViT-B/32  baseline.}
\begin{tabular}{lcccc}
\toprule
Dataset & ImageNette & Caltech-101 & StanfordCars& PathMNIST \\
\midrule
Original & 99.40 & 96.87 & 87.61 & 84.29 \\
Expanded 5$\times$ by SD & 98.51 \green{0.89} & 96.91 \red{0.04} & 90.19 \red{2.58} & 86.81 \red{2.52} \\
Expanded 5$\times$ by DistDiff & 99.30 \green{0.10} & 97.00 \red{0.13} & 91.17 \red{3.56} & 89.47 \red{5.18} \\
\bottomrule
\end{tabular}
\label{table:pretrained}
\vspace{-10pt}
\end{table}



\begin{figure*}[ht]
\begin{center}
\centerline{\includegraphics[width=1.0\linewidth]{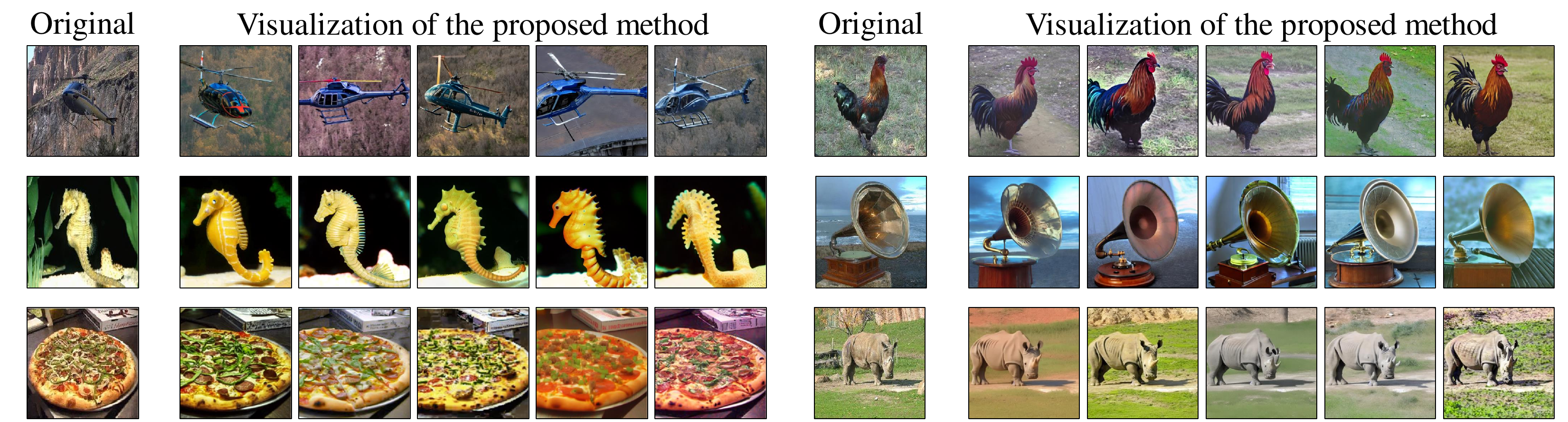}}
\caption{The visualization of synthetic samples generated by our method, showcasing high fidelity, diversity, and alignment with the original data distribution.}
\label{fig:visualization}
\end{center}
\vskip -0.2in
\end{figure*}

\paragraph{Qualitative Analysis} In addition to the quantitative experiment results, we also gain a more intuitive understanding of the diverse changes facilitated by our method through visualization of the generated results. As shown in Figure \ref{fig:visualization}, the images generated using our distribution-aware guidance approach exhibit high fidelity and diverse synthetic changes, including object texture, background, and color contrast. More visualization results can be found in Appendix \ref{append:more_visual_results}.


\subsection{Ablation Study}
\label{sec:ab}
\paragraph{Hierarchical Prototypes} We delved further into how each component of DistDiff impacts its data expansion performance. As depicted in Table \ref{tab:ablation}, utilizing both $\vp_\mathrm{c}$ and $\vp_\mathrm{g}$ contributes to enhancing the model's expansion performance, showcasing their ability to optimize the generated sample distribution at the class-level and group-level, respectively. Moreover, combining $\vp_\mathrm{c}$ and $\vp_\mathrm{g}$ results in a further performance improvement, validating the effectiveness of integrating representations from different hierarchical levels.  Additionally, with the introduction of our approach, the Fréchet Inception Distance (FID) values notably decrease, indicating that our proposed FID effectively optimizes the model to generate samples more aligned with the real distribution, thereby reducing the domain gap between the generated dataset and the real dataset.

\begin{minipage}[t]{0.45\textwidth}
\centering
\captionof{table}{Performance comparison of models trained on original Caltech-101 datasets and 5x expanded datasets by DistDiff.}
\begin{small}
\begin{tabular}{lcc}
\toprule
Backbone       & Original    & DistDiff \\
\midrule
ResNet-50 \cite{resnet}          & 66.71              & 83.09                   \\
ResNeXt-50 \cite{resnext}          & 67.60              & 83.75                   \\
WideResNet-50 \cite{wideresnet}    & 66.51              & 83.51                   \\
MobileNetV2 \cite{mobilenetv2}  & 74.39              & 83.85                   \\
\bottomrule
\end{tabular}
\label{tab:model_comparison} 
\end{small}
\end{minipage}
\hfill
\begin{minipage}[t]{0.45\textwidth}
\centering
\captionof{table}{Comparison of accuracy and FID in expanding Caltech-101 by $5\times$, with and without hierarchical prototypes in DistDiff.}
\begin{small}
\begin{tabular}{cccc}
\toprule
$\vp_\mathrm{c}$ & $\vp_\mathrm{g}$ & Accuracy $\uparrow$  & FID-3K $\downarrow$ \\
\midrule
                &                   & $76.57\pm0.35$ & 72.56  \\
                & \checkmark        & $82.70\pm0.07$ & 68.82  \\
\checkmark      &                   & $82.84\pm0.54$ & 68.66 \\
\checkmark      & \checkmark        & $83.09\pm0.11$ & 67.72 \\
\bottomrule
\end{tabular}
\label{tab:ablation} 
\end{small}
\end{minipage}%

  

\paragraph{Augmentation within Diffusion} In the application of energy guidance, perturbation is introduced at the $M$-th step, and these subsequent predicted data points are optimized. Theoretically, optimizing predictions at different stages yields distinct effects. Our exploration focused on the optimization step $M$, and the experimental results are illustrated in Table \ref{tab:ab_M}. When $M$ is small, indicating optimization at a later stage (i.e., the refinement stage), the change in generated results is already minimal, resulting in relatively consistent optimization outcomes. Conversely, when $M$ increases, corresponding to optimization at an intermediate stage (i.e., the semantic stage), the generated results are in the stage of forming semantics and exhibit significant changes. Hence, this stage plays a crucial role in determining the final generated results. Furthermore, there is a decline in performance during the early chaos stage ($M$=25), as the data points in this initial phase are too chaotic to establish an optimal target for optimization. We observed that achieving higher data expansion performance is possible when optimized in the semantic stage, with optimal results obtained when $M$=20.

\begin{table}[htbp]
\begin{minipage}[b]{0.3\linewidth}
\centering
\captionof{table}{Comparison of optimization in different phases.}
\begin{tabular}{lc}
\toprule
$M$ & Accuracy \\
\midrule
1 & $81.54 \pm 0.32$ \\
10 & $82.21 \pm 0.22$ \\
20 & $82.36 \pm 0.05$ \\
25 & $82.11 \pm 0.55$ \\
\bottomrule
\end{tabular}
\label{tab:ab_M}
\end{minipage}%
\hfill
\begin{minipage}[b]{0.3\linewidth}
\centering
\captionof{table}{Ablation of the number $K$ of $\vp_\mathrm{g}$ in DistDiff.}
\begin{tabular}{lc}
\toprule
$K$ & Accuracy \\
\midrule
2 & $82.69 \pm 0.51$ \\
3 & $83.09 \pm 0.11$ \\
4 & $83.08 \pm 0.13$ \\
5 & $83.08 \pm 0.21$ \\
\bottomrule
\end{tabular}
\label{tab:ab_K}
\end{minipage}%
\hfill
\begin{minipage}[b]{0.3\linewidth}
\centering
\captionof{table}{Results of introducing more optimization steps.}
\begin{tabular}{lc}
\toprule
Step & Accuracy \\
\midrule
1 & $82.54 \pm 0.54$ \\
2 & $82.94 \pm 0.43$ \\
3 & $82.77 \pm 0.19$ \\
4 & $82.55 \pm 0.30$ \\
\bottomrule
\end{tabular}
\label{tab:ab_P}
\end{minipage}
\end{table}

\paragraph{More Optimization Steps}
Furthermore, a natural idea arises regarding the potential improvement in effectiveness through the optimization of more optimization steps. Therefore, we further explored increasing the number of steps in the semantic stages.
As shown in Table \ref{tab:ab_P}, increasing the number of optimization steps in semantic stages enhances performance. However, further increases in optimization steps lead to a decline in performance. This can be attributed to excessive optimization strength in energy guidance, which causes data distortion.

\paragraph{Compared with Direct Guidance on Latent Point} 
We evaluate our transform guidance strategy against an alternative strategy that directly guides the latent data points while ignoring the residual transform preprocess, a method found useful in previous works \cite{ho2022classifier, yu2023freedom}. 
We initially conducted a grid search for this alternative strategy to find the optimal learning rate, $\rho$ (i.e., [0.1, 1, 10, 20]), and guide step, $M$ (i.e., [1, 10, 20]). We found the best result of 76.77\% was achieved with $\rho = 10$  and $M = 10$, which is only slightly better than the original Stable Diffusion but lags behind our DistDiff, which achieved 83.19\%. This indicates that applying the residual multiplicative transformation to the latent feature offers more optimization potential.

\begin{figure*}[htbp]
\centering
\begin{minipage}[b]{0.24\textwidth}
    \centering
    \includegraphics[width=\textwidth]{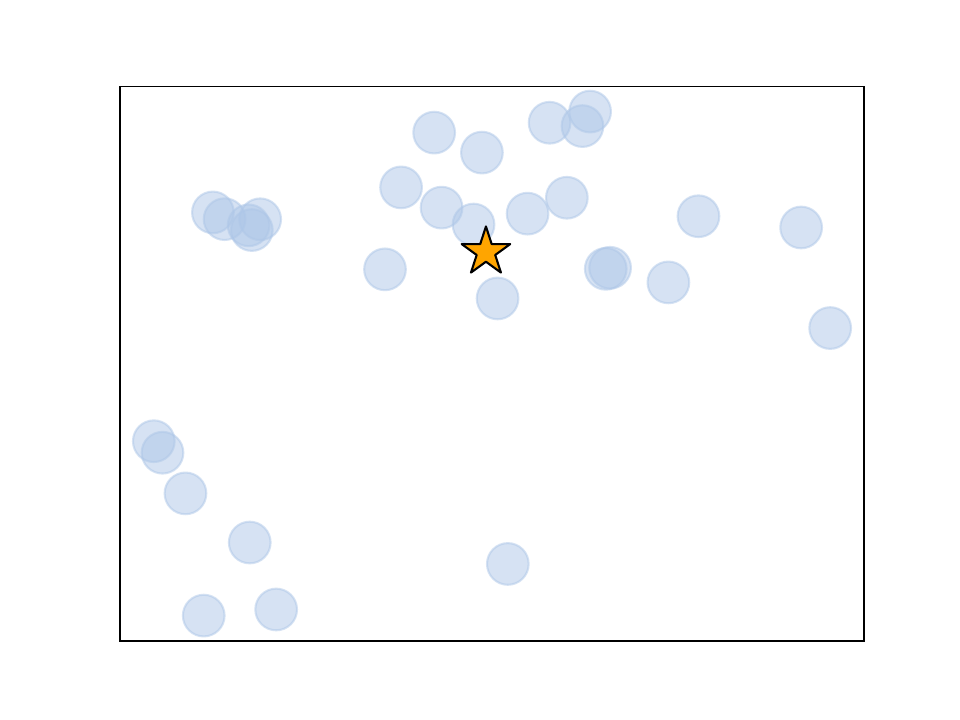}
    \small{(a) $K=1$}
\end{minipage}
\begin{minipage}[b]{0.24\textwidth}
    \centering
    \includegraphics[width=\textwidth]{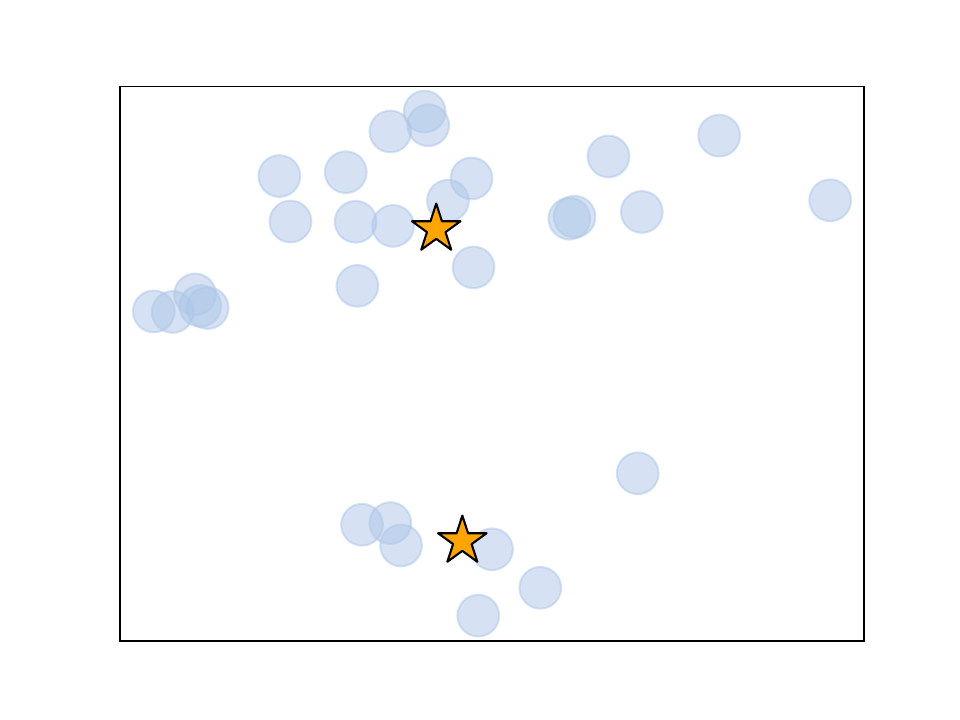}
     \small{(b) $K=2$}
\end{minipage}
\begin{minipage}[b]{0.24\textwidth}
    \centering
    \includegraphics[width=\textwidth]{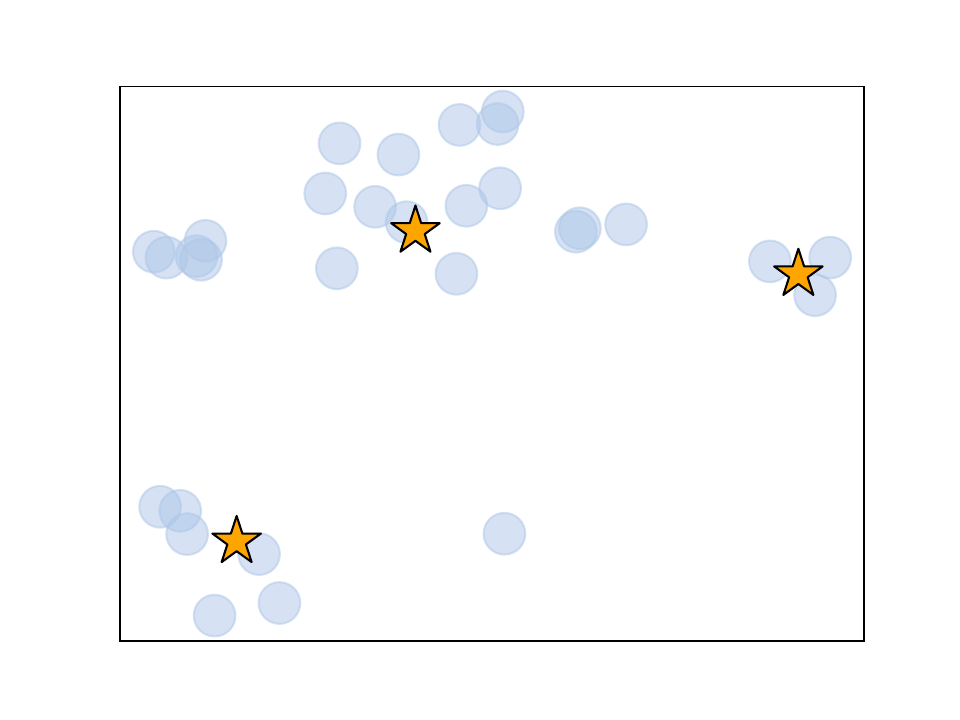}
     \small{(c) $K=3$}
\end{minipage}
\begin{minipage}[b]{0.24\textwidth}
    \centering
    \includegraphics[width=\textwidth]{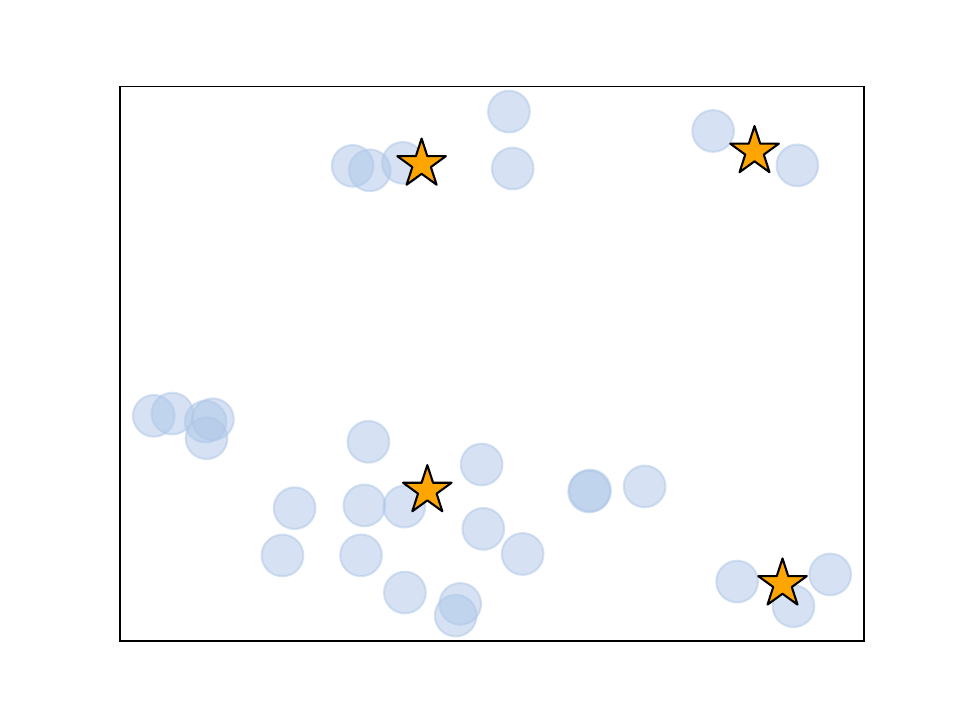}
     \small{(d) $K=4$}
\end{minipage}
\caption{The visualization of group-level prototypes alongside original sample features. Here \textbf{\textcolor{blue}{$\bullet$}}  is the sample point and \textbf{\textcolor{orange}{$\star$}} is group-level prototype.  By selecting an appropriate number $K$, these prototypes effectively span the feature space, providing an approximation of the real data distribution.}
\label{fig:ab_k}
\vspace{-10pt}
\end{figure*}

\paragraph{Determination of Group-Level Prototype Number $K$} The determination of the number $K$ of group-level prototypes is crucial for accurately approximating the real data distribution. In Table \ref{tab:ab_K}, we compare the outcomes associated with varying numbers of prototypes. The results highlight that the optimal number of prototypes is found at $K=3$. We posit that an insufficient number of prototypes may impede the characterization of the real distribution, leading to diminished performance. Conversely, an excessive number of prototypes may lead to overfitting of noisy sample points, also resulting in suboptimal performance. Furthermore, we present a visualization analysis of group-level prototypes in Figure \ref{fig:ab_k}. The visual representation demonstrates that an appropriate number of group-level prototypes can effectively cover the real distribution space, aligning with the underlying motivation of our DistDiff.

\paragraph{Computational Efficiency} 
Our DistDiff is not only training-free but also highly efficient in processing. As illustrated in Figure \ref{fig:overview}, DistDiff introduces only a few optimization steps in the original diffusion process. We analyze the time costs of our methods. Stable Diffusion generates per sample in 12.65 seconds on average, while our DistDiff achieves the same in 13.13 seconds, with the increased time costs being negligible. We can conclude that DistDiff achieves a notable improvement over stable diffusion models, with only a slight increase in computation costs.

\section{Conclusion}
\label{sec:conclusion}
\vspace{-10pt}
This paper presents DistDiff, a distribution-aware data expansion method employing a stable diffusion model for data expansion. The proposed method optimizes the diffusion process to align the synthesized data distribution with the real data distribution. Specifically, DistDiff constructs hierarchical prototypes to effectively represent the real data distribution and refines intermediate features within the sampling process using energy guidance. We evaluate our method through extensive experiments on six datasets, showcasing its superior performance over existing methods.
\paragraph{Limitations and Societal Impacts} 
Although DistDiff achieves better FID scores and enhances classifier performance without the need for training, it may incur additional computation time, which accumulates as data scales up. Additionally, our method requires an extra guide model, increasing development costs. Integrating a fast sampler \cite{luo2023latent} and lightweight guide model represents a promising direction for maximizing the effectiveness of diffusion-based data expansion methods in classifier training. Furthermore, adopting this approach in real-life applications requires careful consideration of its potential impact. We delve into the societal impact of employing these models in Appendix \ref{append:social_impact}.


{\small
\bibliographystyle{ieee}
\bibliography{reference}
}

\newpage
\appendix
\section{Social Impact} 
\label{append:social_impact}
Generative models \cite{DALLE2, stablediffusion} can significantly reduce the costs associated with manual data collection and annotation. Our approach builds an efficient method that enhances the distributional consistency of these generative models for downstream tasks without requiring training, thereby improving performance on downstream classifiers. This capability can assist organizations and researchers with limited data resources in developing more effective machine learning models. 

However, since generative models are pre-trained on vast, diverse vision-language datasets from the internet, these data may contain social biases and stereotypes \cite{cho2023dall, naik2023social, ross2020measuring}, leading to discriminatory generated outputs. Therefore, integrating mechanisms to detect and mitigate biases is crucial. 
Nevertheless, our method guides the generation process decisions based on task-specific hierarchical prototypes, fostering AI systems that better align with downstream task distributions and exhibit fewer biases. 

Another potential concern is the misuse of generated data, which could be exploited for malicious purposes such as deepfakes \cite{lyu2020deepfake}, resulting in misinformation dissemination and adverse societal impacts. Proper constraints on the proliferation and correct usage of generative models are crucial. This necessitates the establishment of relevant regulations and guidelines to ensure the responsible development and utilization of synthetic data models.

\section{More Implementation Details}
\subsection{Datasets}
\label{append:dataset}
Table \ref{tab:dataset_summary} provides the detailed statistics of six experimental datasets, including Caltech-101 \cite{caltech101}, CIFAR100-Subset \cite{cifar}, StandardCars \cite{car}, ImageNette \cite{imagenette}, DTD \cite{dtd}, and \pathmnist \cite{yang2023medmnist}. Cifar100-Subset and PathMNIST are subsets that randomly sampled from the original trainset with 100 samples per class. 
Specifically, our evaluation encompasses datasets that span diverse scenarios, incorporating generic objects, fine-grained categories, and textural images. These diverse datasets enable a comprehensive assessment of both the robustness and effectiveness of our proposed methods. Additionally, we also provide the text template employed in the conditional diffusion process in Table \ref{tab:dataset_temp}.

\begin{table*}[h]
\caption{Summary of our six experimental datasets.}
\label{tab:dataset_summary}
\vskip 0.15in
\begin{center}
\begin{small}
\begin{sc}
\begin{tabular}{lccc}
    \toprule
    \textbf{Name} & \textbf{Classes} & \textbf{Size (Train / Test)} & \textbf{Description}  \\
    \midrule
    Caltech-101      & 100    & 3000 / 6085   & Recognition of generic objects     \\ 
    CIFAR100-Subset  & 100  & 10000 / 10000   & Recognition of generic objects   \\
    StandardCars    & 196 & 8144 / 8041 & Fine-grained classification of cars     \\ 
    ImageNette      & 10 & 9469 / 3925  & Recognition of generic objects \\
    DTD             & 47 & 3760 / 1880 & Texture classification  \\  
    PathMNIST       &  9 &  900 / 7180   & Recognition of colon pathology image \\
    \bottomrule
  \end{tabular}
\end{sc}
\end{small}
\end{center}
\vskip -0.1in
\end{table*}

\begin{table*}[h]
\caption{Text templates for six experimental datasets.}
\label{tab:dataset_temp}
\vskip 0.15in
\begin{center}
\begin{small}
\begin{sc}
\begin{tabular}{lc}
    \toprule
    \textbf{Name} & \textbf{Template} \\
    \midrule
    Caltech-101       & “a photo of a [class].”   \\ 
    CIFAR100-Subset   & “a photo of a [class].”  \\
    StandardCars      &  “A photo of a [class] car.”  \\ 
    ImageNette        & “a photo of a [class].” \\
    DTD             & “[class] texture.”  \\  %
    PathMNIST       &  “a colon pathological image of [class].” \\
    \bottomrule
  \end{tabular}
\end{sc}
\end{small}
\end{center}
\vskip -0.1in
\end{table*}

\subsection{Our Method}
\label{append:implement}
We implement our methods using the PyTorch framework with Python 3.10.6. We utilize the diffuser \cite{diffusers} to implement our base diffusion models. For all datasets, we generate images with a noise strength of 0.5, except for the medical image dataset PathMNIST, where we use a noise strength of 0.2. The guidance step $M$ in our method is 20 by default, except for 10 for PathMNIST. The half-precision floating-point is used in the generation process to reduce memory costs.
The generation, training, and evaluation processes are conducted on a single GeForce RTX 3090 GPU. We report the best test accuracy averaged over three runs and calculate the FID metric using 3000 samples. In order to reduce generation time costs, we pre-compute and save the latent embeddings for all training samples. We use Stable Diffusion 1.4 as our base generation model\footnote{We use the pretrained weights ``CompVis/stable-diffusion-v1-4'' from Hugging Face. \url{https://huggingface.co/CompVis/stable-diffusion-v1-4}}. 

\subsection{Synthesis-Based Augmentation Contenders}
\label{append:Synthesis-details}
Due to benchmark and training setting differences in existing works \cite{gifsd, azizi2023synthetic, li2022bigdatasetgan}, in order to fairly compare existing methods with ours, we reproduced two state-of-the-art generation-based methods: GIF-SD \cite{gifsd} and LECF \cite{azizi2023synthetic} on our benchmarks and base diffusion model. For GIF-SD, we used a pretrained CLIP-ViT-B32 model to facilitate its class-maintained optimization, and a batch size of 4 to facilitate its KL-divergence based diverse sampling. For LECF, we generated 200 prompts for each class prompt, and generated samples with sizes equal to our method's, filtering these samples with a specified threshold using a pretrained ResNet50-CLIP model. As shown in Section \ref{sec:main_results}, our methods surpass these contenders.

\subsection{Transformation-Based Augmentation Contenders}
\label{append:Transformation-details}
We outline the transformation-based augmentation methods compared in our experiments. 
The effectiveness of all these methods are evaluated on the same datasets and with the same configuration (e.g., learning rate, epochs, batch size, etc.) as our method.

\textbf{Default}: We use random crop, random horizontal flip, and random rotation with a 15-degree angle as default augmentation strategies. 

\textbf{AutoAug \cite{cubuk2018autoaugment}}: We employ the AutoAugment function in PyTorch with the widely used ImageNet policy, which includes a diverse range of color and shape transformations. During training, one transformation strategy is randomly selected and applied.

\textbf{RandAug \cite{cubuk2020randaugment}}: Similar to AutoAug, we randomly choose two operations from a predefined policy list and apply them to the training samples.

\textbf{Random Erasing \cite{cutout}}: We use the random erase function in PyTorch for Random Erasing. This function randomly selects a rectangular region within an image and erases its pixels with a 50\% probability. The proportion of erased area relative to the input image ranges from 0.02 to 0.33, and the aspect ratio of the erased area ranges from 0.3 to 3.3.

\textbf{GridMask \cite{chen2020gridmask}}: GridMask is implemented with its official configuration. The probability of applying GridMask linearly increases from 0 to 0.8 as training epochs progress up to the 80th epoch, after which it remains constant until 100 epochs.

\textbf{MixUp \cite{zhang2017mixup}}: Synthetic interpolated images are generated within each data batch. We sample the interpolation strength from a beta distribution (beta = 1). The loss function is also adjusted accordingly.

\textbf{CutMix \cite{yun2019cutmix}}: Similar to MixUp, CutMix replaces specified regions in original images with input from another image. The loss function is modified accordingly.

\section{Pseudocode}
\label{append:pseudocode}
We present the pseudocode for our algorithm in Algorithm \ref{algorithm:Sampling}, illustrating the hierarchical energy guidance within the diffusion sampling process.

\begin{algorithm}[ht]
	\caption{Optimization Process of our proposed DistDiff}
	\label{algorithm:Sampling}
	\begin{algorithmic}
		\STATE {\bfseries Input:} Hierarchical prototypes $\vp_\mathrm{c}$ and $\vp_\mathrm{g}$; Data point $\vz_T$; Optimization step $M$; Pre-defined parameters $\beta_t$; Perturbation constraint $\epsilon$; Pre-trained feature extractor $\vtheta$; Denoising network $\psi$.
            \FOR{$t = T-1, \dots, 0$}
                \STATE $\delta\sim\gN(\mathbf{0}, \mathbf{I})$ if $t > 0$, else $\delta=0$.
                \STATE $\vz_{t} = (1+\frac{1}{2}\beta_{t+1})\vz_{t+1} + \beta_{t+1}\psi(\vz_{t+1},t+1)+\sqrt{\beta_{t+1}}\delta$
                \IF{$t = M$}
                    \STATE Initialize $\ve\sim\gU(0,1)$, $\vb\sim\gN(0,1)$.
                    \STATE $\vz_t = (1+\ve)\vz_t+\vb$
                    \STATE $\vz_{0|t} = (\vz_t-\sqrt{1-\alpha_t}\psi(\vz_t,t)) / \sqrt{\alpha_t}$
                    \STATE $\gD_\vtheta^\mathrm{c}(\vz_{0|t}, \vp_\mathrm{c}) = \lVert \vtheta(\vz_{0|t}) - \vp_\mathrm{c} \rVert_2$
                    \STATE $\mathcal{D}_\vtheta^\mathrm{g}(\vz_{0|t}, \vp_\mathrm{g}) = \lVert \vtheta(\vz_{0|t}) - \vp_\mathrm{g} \rVert_2$
                    \STATE $\vg_t = \gD_\vtheta^\mathrm{c}(\vz_{0|t}, \vp_\mathrm{c}) + \mathcal{D}_\vtheta^\mathrm{g}(\vz_{0|t}, \vp_\mathrm{g})$
                    \STATE update $\ve', \vb' \gets {\arg\min}_{e,b} \vg_t$ (Equation \ref{Eq:final_guidance})
                    \STATE update $\vz_{t} \gets \vz'_{t} = \gP_{z,\epsilon}((1+\ve')\vz_t+\vb')$
                \ENDIF
            \ENDFOR
	\STATE {\bfseries Output:} $\vz_{0}$.
	\end{algorithmic}
\end{algorithm}


\section{More Experimental Results}
\subsection{Model Performance}
\paragraph{DistDiff is Robust to Guidance Model}
As mentioned in Section \ref{sec:hierarchical_design}, we employ a extra feature extractor as our guidance model. To assess the impact of different guidance models, we compared two backbones: a ResNet-50 \cite{resnet} trained on the original dataset from scratch (weak backbone) and CLIP \cite{Cherti_2023_CVPR}, a strong backbone pre-trained on large datasets and fine-tuned on the original dataset. Table \ref{tab:guidance_model_comparison} presents the accuracy of the guidance models and the corresponding tuned classifier (ResNet-50). Our DistDiff method demonstrates robustness, showing negligible changes in accuracy (0.25\%) across different guidance models. This highlights the robustness of our approach. To ensure fairness in comparison without knowledge leakage from large pre-trained models, we default to using a randomly initialized ResNet-50 trained on the original dataset as our guidance model.

\begin{table}[htbp]
    \centering
    \caption{Comparison of guidance models on Caltech-101 dataset. We compared the accuracy of two guidance models on the original Caltech-101 dataset. Additionally, we evaluated the performance of a downstream classifier trained on the $5\times$ expanded dataset using corresponding guide model.}
    \begin{tabular}{lcc}
        \toprule
        \textbf{Guide Model} & \multicolumn{2}{c}{\textbf{Accuracy (\%)}} \\
        \cmidrule(lr){2-3}
         & \textbf{Guide Model} & \textbf{Downstream Classifier} \\
        \midrule
        Weak (Random initialized and trained ResNet) & $66.71 \pm 0.47$ & $82.94 \pm 0.43$ \\
        Strong (Pre-trained and finetuned CLIP) & $92.24 \pm 0.25$ & $83.19 \pm 0.69$ \\
        \bottomrule
    \end{tabular}
    \label{tab:guidance_model_comparison}
\end{table}

\paragraph{Determination of Guidance Scale $\rho$.} 
We compared different learning rates $\rho$, and the experimental results are shown in Table \ref{tab:rho_compare}. A learning rate that is too low results in insufficient optimization. Conversely, a learning rate that is too high causes over-optimization, leading to image distortion. Both result in suboptimal performance. We used $\rho=10$, which achieves the best results, as the default learning rate.

\begin{table}[htbp]
    \centering
    \caption{Comparison of different learning rate $\rho$.}
    \begin{tabular}{lcccc}
        \toprule
        $\rho$ & 0.1 & 1.0 & 10.0 & 20.0 \\
        \midrule
        Accuracy (\%) & $82.49 \pm 0.33$ & $82.74 \pm 0.32$ & $83.09 \pm 0.11 $ & $82.46 \pm 0.35$ \\
        \bottomrule
    \end{tabular}
    \label{tab:rho_compare}
\end{table}

\paragraph{Comparison of Varying Gradient Weight.} 
We compared different contribution levels of two hierarchical prototypes. Specifically, we scaled the gradient of $p_g$ by a certain coefficient, $\lambda_g$. The results in Table \ref{tab:loss_weight} indicate that an appropriate scaling weight can further enhance overall performance.

\begin{table}[htbp]
    \centering
    \caption{Comparison of different gradient weights $\lambda_g$.}
    \begin{tabular}{lccccccc}
        \toprule
        $\lambda_g$ & 0.1 & 0.3 & 0.5 & 0.7 & 0.9 & 1.0 & 2.0     \\
        \midrule
        Accuracy (\%) & $82.61$ & $82.79$ & $83.14$ & $83.12$ & \textbf{83.38} & $83.09$ & $82.73$ \\
        \bottomrule
    \end{tabular}
    \label{tab:loss_weight}
\end{table}

\subsection{Further Analysis}
\label{append:further_analysis}
\begin{figure*}[htbp]
\centering
\centering
\includegraphics[width=0.55\textwidth]{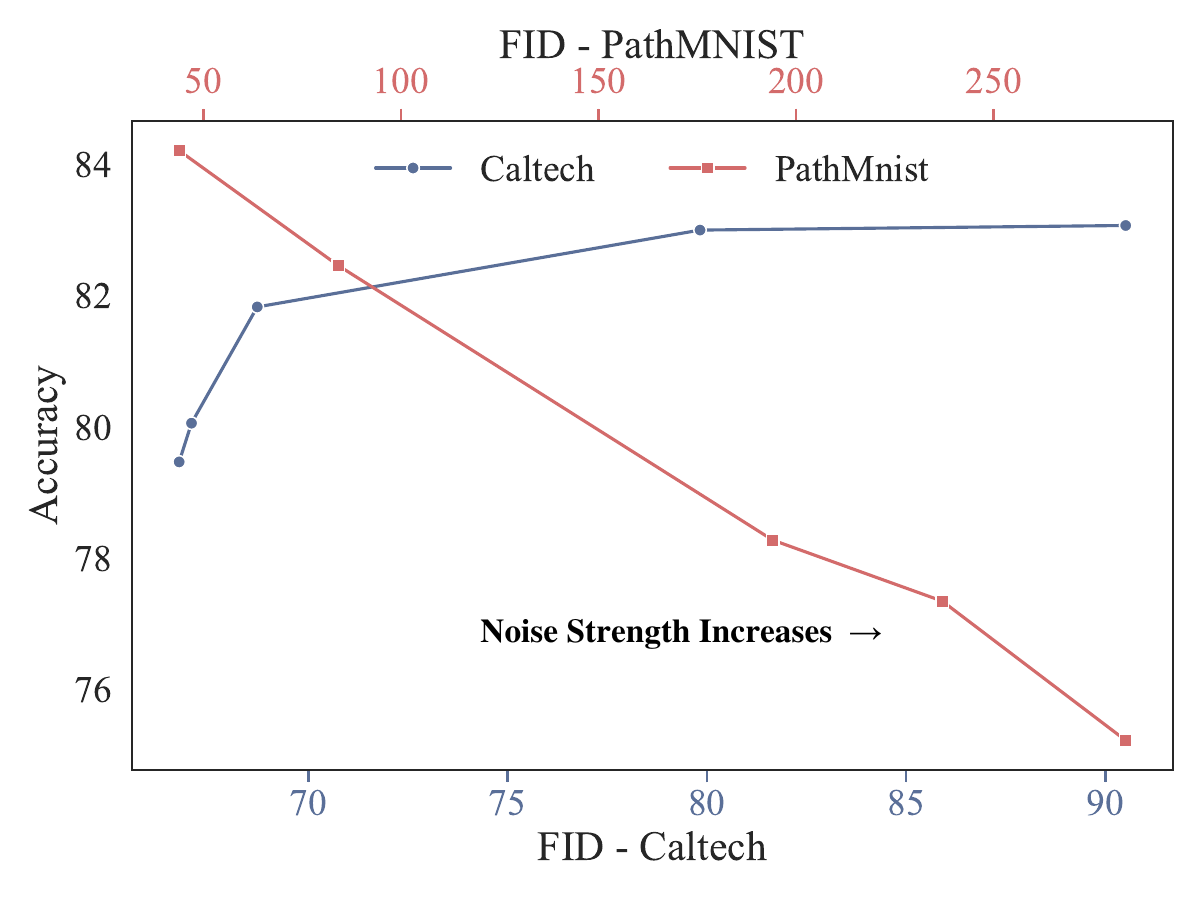}
\caption{Comparison with FID and accuracy across varying noise strengths.}
\label{fig:append_fid_accuracy}
\vspace{-10pt}
\end{figure*}

\paragraph{Trade-Off Between Fidelity and Diversity}
The data expansion task requires both high fidelity and diversity for effective model training. However, this principle does not universally apply across all scenarios. We assessed the trade-off between high fidelity and diversity by adjusting the diffusion model's strength in adding noise to original images. As the noise strength increases, the diversity of generated data enhances, accompanied by a decrease in FID scores. The resulting accuracy and FID scores are presented in Figure \ref{fig:append_fid_accuracy}, where we observed an inverse relationship between FID (fidelity metric) and accuracy across the Caltech-101 and PathMNIST datasets.
We found that introducing more new content on general datasets, such as Caltech-101, can benefit model training by emphasizing the need for diversity. In contrast, on medical datasets with substantial distribution shifts, maintaining the original data distribution and minimizing disturbances proves to be more effective.

\subsection{More Visualization Results}
\label{append:more_visual_results}
\begin{figure*}[htbp]
\centering
\centering
\includegraphics[width=0.8\textwidth]{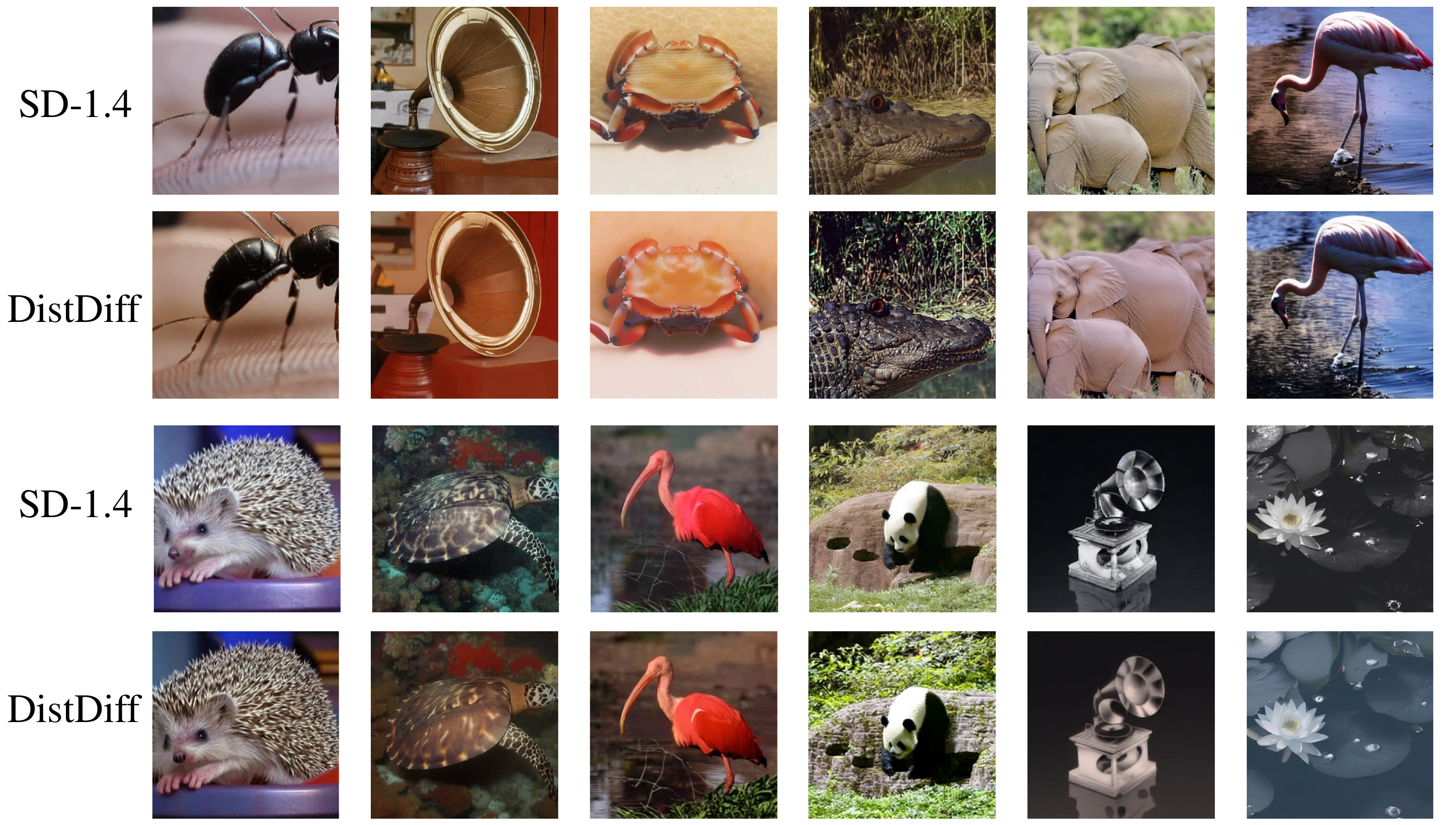}
\caption{Comparison of visualizations between original Stable Diffusion 1.4 and our DistDiff.}
\label{fig:compared_sd}
\vspace{-10pt}
\end{figure*}

In this section, we provide visualization comparison between original stable diffusion (SD) and our DistDiff in Figure \ref{fig:compared_sd}. Visualization results indicate that our method exhibits finer texture details, more diverse style variations, and stronger foreground-background contrasts, making our samples align real sample distributions. However, the main content of our images still shows minimal deviation from SD, making it challenging for users to discern differences through visual observation alone. In Section \ref{sec:ab}, we provide quantitative analysis to validate that our method generates data with more consistent distributions, thereby enhancing performance in downstream classification tasks. 

\begin{figure*}[htbp]
\centering
\begin{minipage}[b]{0.44\textwidth}
    \centering
    \includegraphics[width=\textwidth]{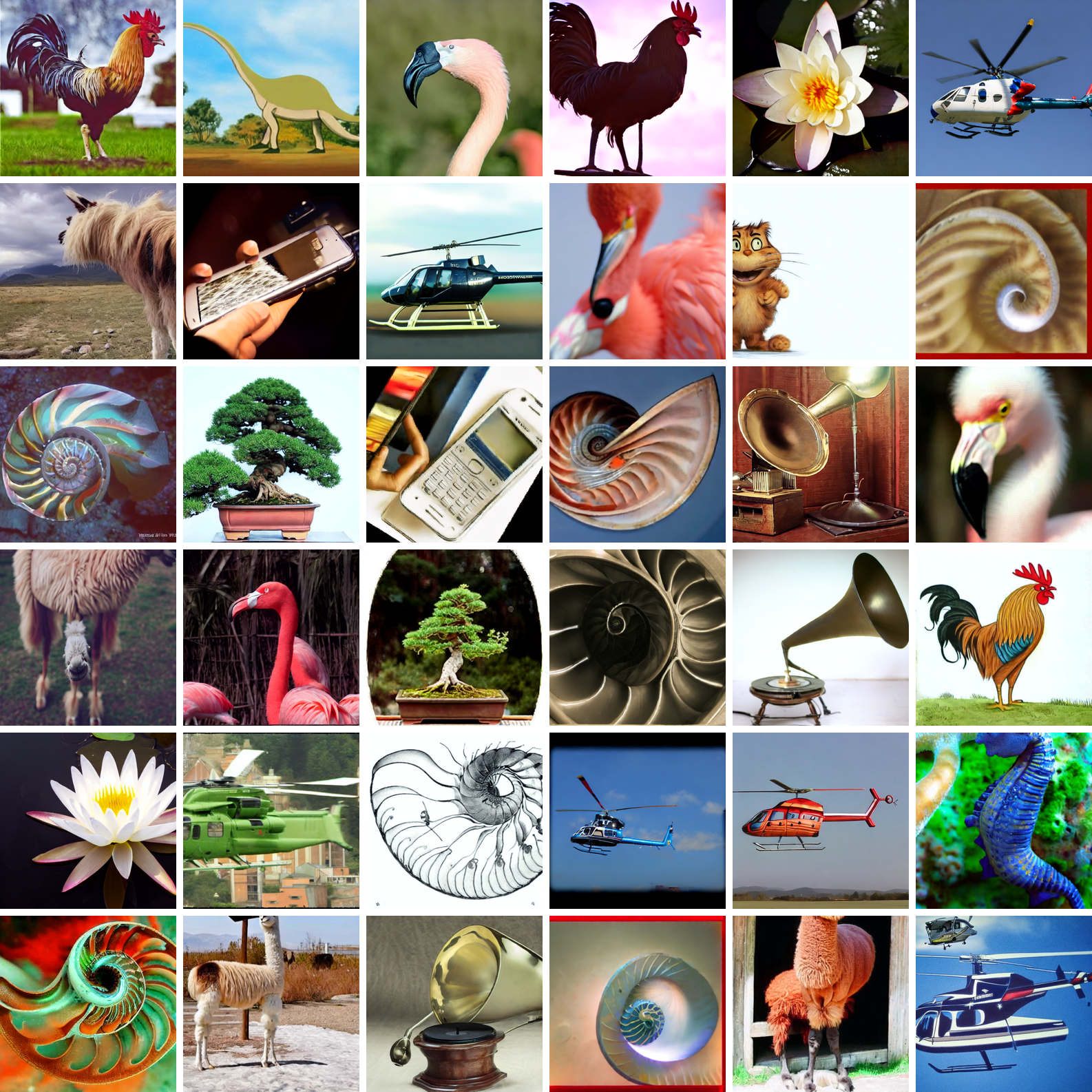}
    \small{Caltech-101}
\end{minipage}
\hfill
\begin{minipage}[b]{0.44\textwidth}
    \centering
    \includegraphics[width=\textwidth]{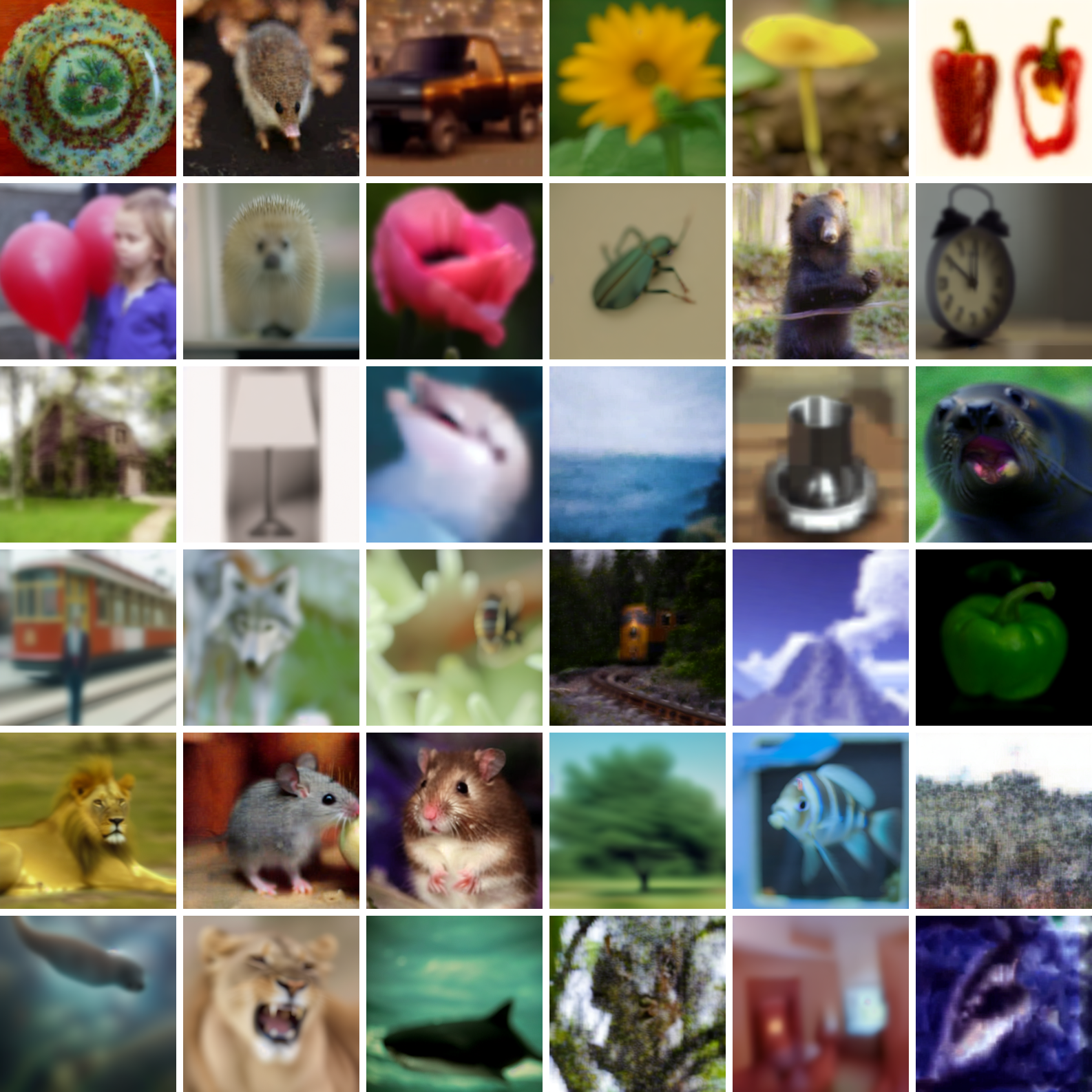}
     \small{Cifar100-Subset}
\end{minipage}
\begin{minipage}[b]{0.44\textwidth}
    \centering
    \includegraphics[width=\textwidth]{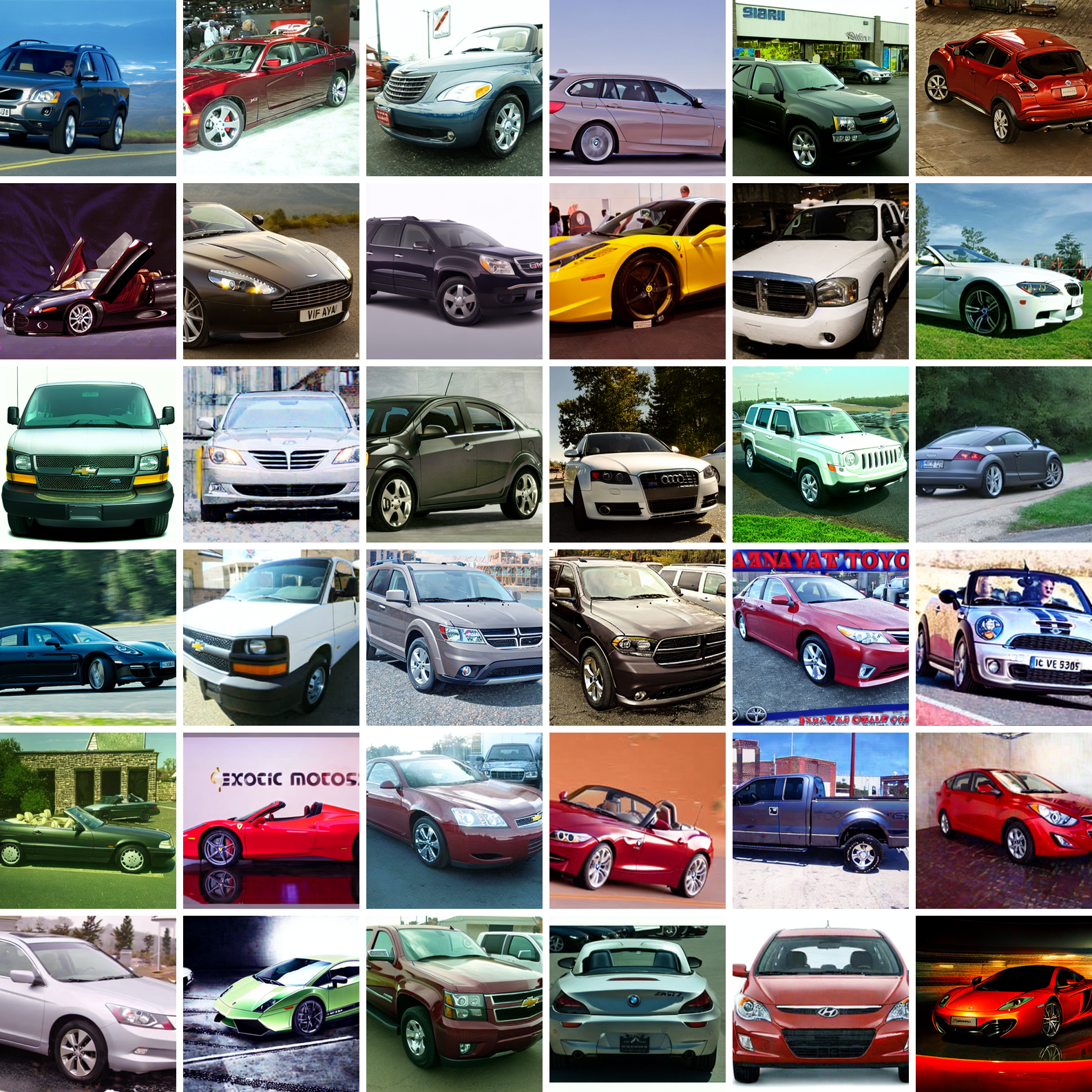}
    \small{StanfordCars}
\end{minipage}
\hfill
\begin{minipage}[b]{0.44\textwidth}
    \centering
    \includegraphics[width=\textwidth]{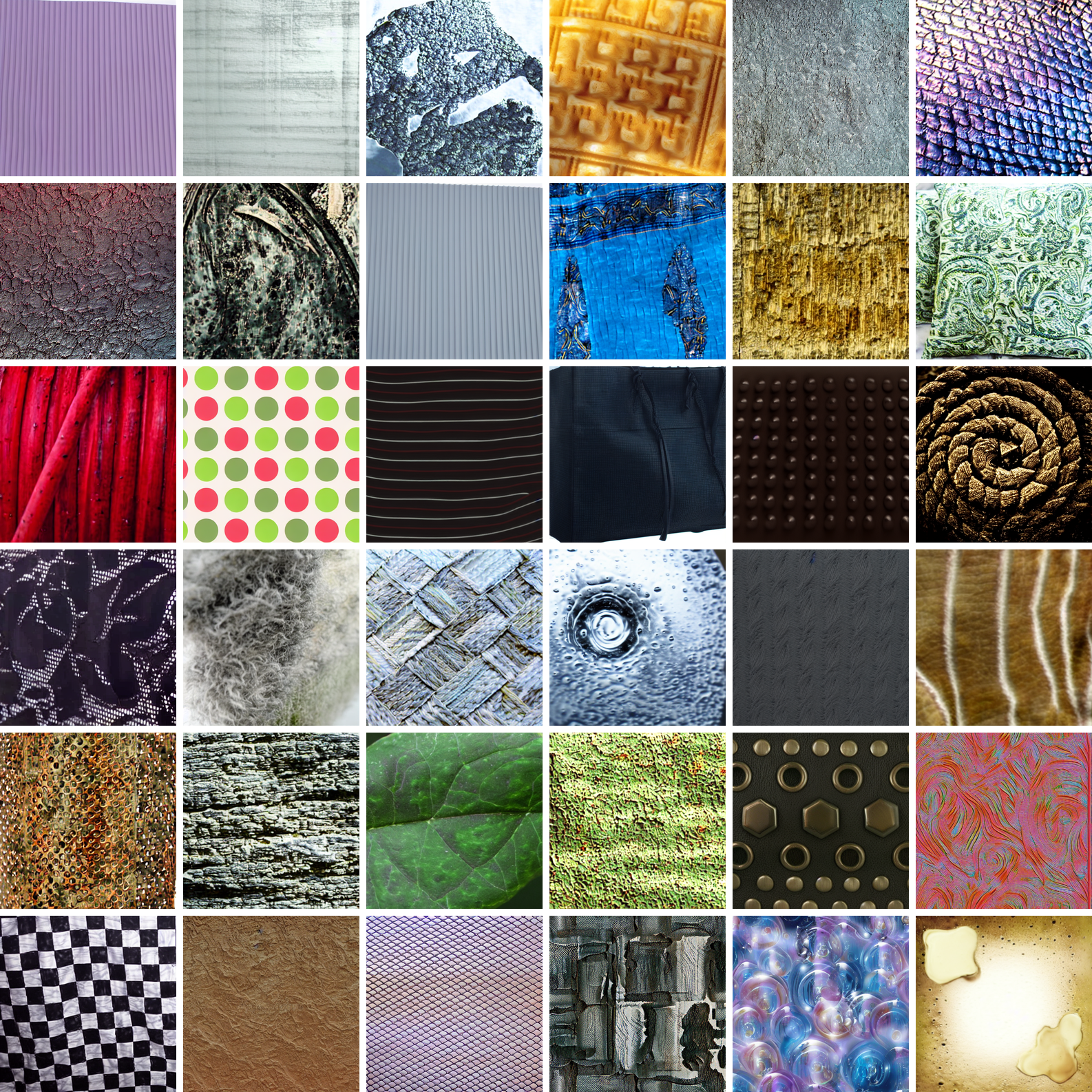}
     \small{DTD}
\end{minipage}
\begin{minipage}[b]{0.45\textwidth}
    \centering
    \includegraphics[width=\textwidth]{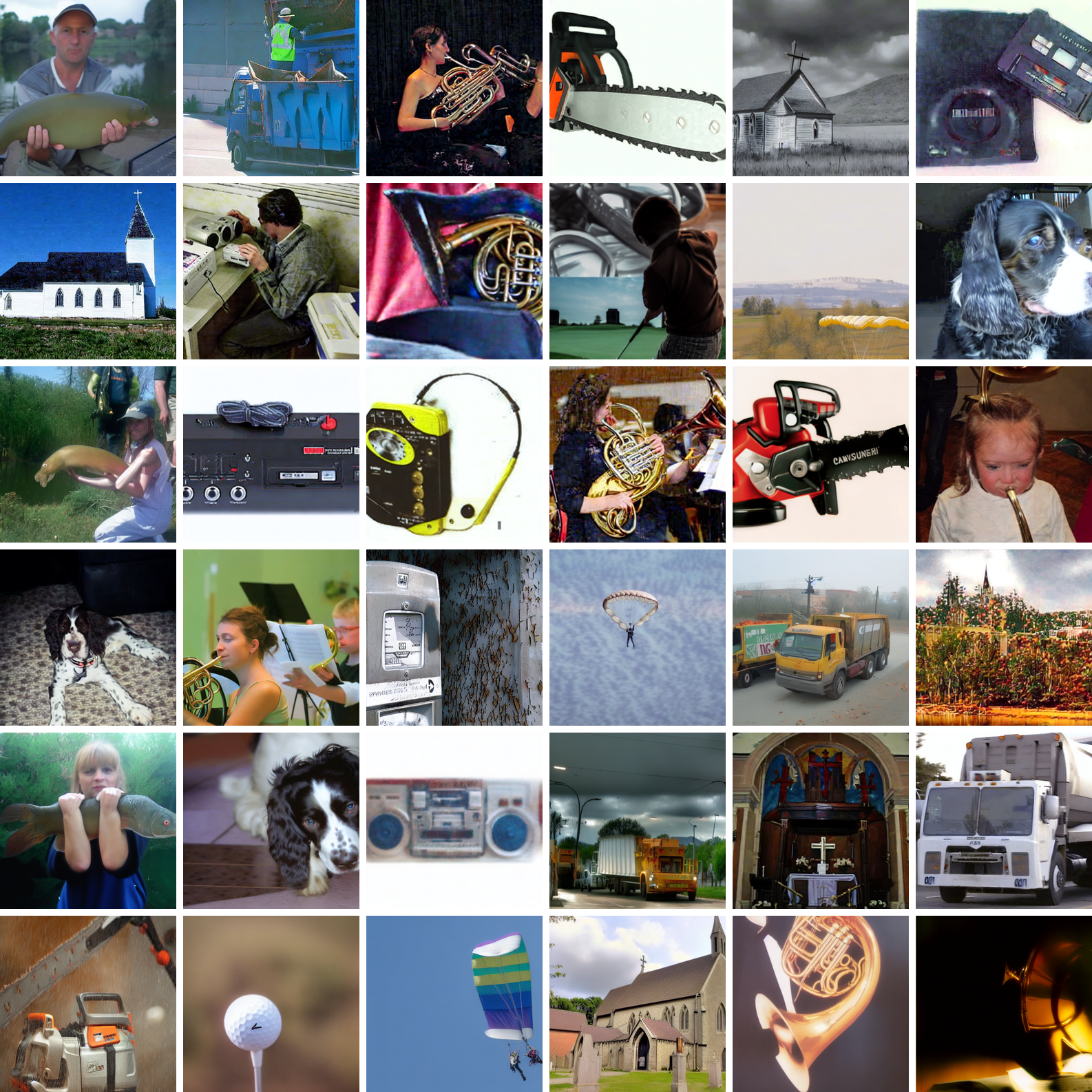}
    \small{StanfordCars}
\end{minipage}
\hfill
\begin{minipage}[b]{0.45\textwidth}
    \centering
    \includegraphics[width=\textwidth]{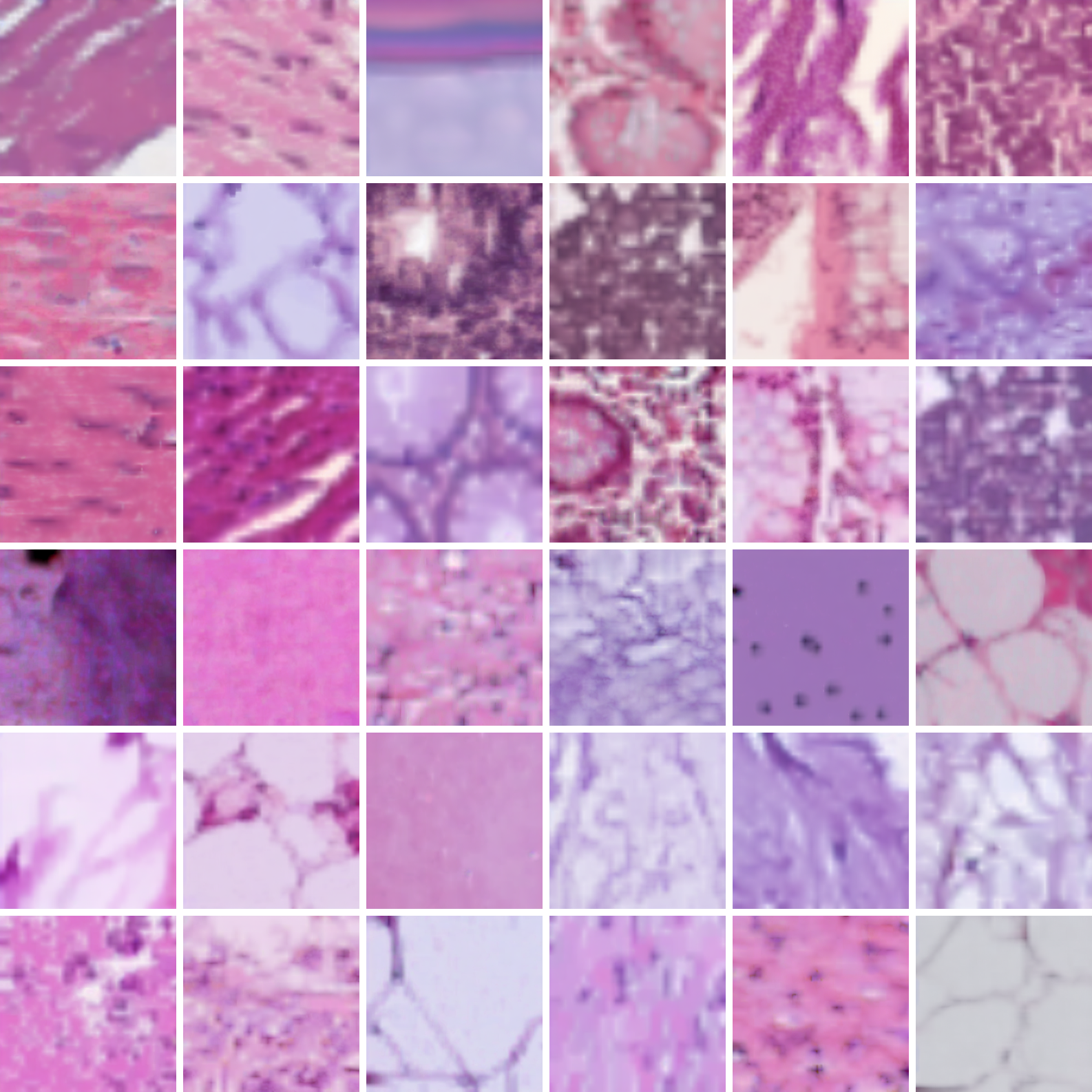}
     \small{PathMNIST}
\end{minipage}
\caption{Visualization of synthetic images produced by our method.}
\label{fig:more_visual}
\end{figure*}

In addition, we visualize more generated samples across six datasets in Figure \ref{fig:more_visual} to demonstrate the effectiveness of DistDiff. These visualizations confirm that our approach can generate samples with more distribution-consistent patterns.

\end{document}